\documentclass[journal]{IEEEtai}

\usepackage[colorlinks,urlcolor=blue,linkcolor=blue,citecolor=blue]{hyperref}
\usepackage{color,array}
\usepackage{graphicx}

\usepackage{cite}
\usepackage{dsfont}
\usepackage{xtab}
\usepackage{caption}
\usepackage{array}
\usepackage{makecell}
\usepackage{upgreek}
\usepackage[ruled,vlined]{algorithm2e}
\usepackage{amsmath,amssymb,amsfonts}
\usepackage{algorithmic}
\usepackage{textcomp}
\usepackage{xcolor}
\usepackage[margin=0.75in]{geometry}
\usepackage{multirow}
\usepackage[normalem]{ulem}

\DeclareMathOperator*{\argmax}{arg\,max}
\DeclareMathOperator{\EX}{\mathbb{E}}

\begin{document}

\title{Hierarchical Reinforcement Learning for Air Combat at DARPA's AlphaDogfight Trials} 

\author{Adrian P. Pope*, Jaime S. Ide*, Daria Mi\'{c}ovi\'{c}, Henry Diaz, Jason C. Twedt, Kevin Alcedo,\\
Thayne T. Walker, David Rosenbluth, Lee Ritholtz, and Daniel Javorsek II

\thanks{* These authors contributed equally to this work.}
\thanks{Adrian P. Pope and Lee Ritholtz are with Primordial Labs, and at the time of the work were with the Lockheed Martin Artificial Intelligence Center, Applied AI Team. (e-mail: adrian@primordial-labs.com).}
\thanks{Jaime S. Ide, Daria Mi\'{c}ovi\'{c}, Henry Diaz, David Rosenbluth, Jason C. Twedt, Thayne T. Walker, and Kevin Alcedo are with the Lockheed Martin Artificial Intelligence Center, Applied AI Team. (e-mail: jaime.s.ide@lmco.com).}
\thanks{Daniel Javorsek II is with the United States Air Force.}
\thanks{This article is based upon work sponsored by the Defense Advanced Research Projects Agency (DARPA).}
\thanks{This work has been accepted for publication in \textit{IEEE Transactions on Artificial Intelligence}. Published version: vol. 4, no. 6, pp. 1371--1385, 2023, doi: \href{https://doi.org/10.1109/TAI.2022.3222143}{10.1109/TAI.2022.3222143}.}
\thanks{\copyright~2022 IEEE. Personal use of this material is permitted. Permission from IEEE must be obtained for all other uses, in any current or future media, including reprinting/republishing this material for advertising or promotional purposes, creating new collective works, for resale or redistribution to servers or lists, or reuse of any copyrighted component of this work in other works.}
}

\markboth{}
{Pope and Ide \MakeLowercase{\textit{et al.}}: Hierarchical Reinforcement Learning for Air Combat at DARPA's AlphaDogfight Trials}

\maketitle

\begin{abstract}
Autonomous control in high-dimensional, continuous state spaces is a persistent and important challenge in the fields of robotics and artificial intelligence. Because of high risk and complexity, the adoption of AI for autonomous combat systems has been a long-standing difficulty. In order to address these issues, DARPA's AlphaDogfight Trials (ADT) program sought to vet the feasibility of and increase trust in AI for autonomously piloting an F-16 in simulated air-to-air combat. Our submission to ADT solves the high-dimensional, continuous control problem using a novel hierarchical deep reinforcement learning approach consisting of a high-level policy selector and a set of separately trained low-level policies specialized for excelling in specific regions of the state space. Both levels of the hierarchy are trained using off-policy, maximum entropy methods with expert knowledge integrated through reward shaping. Our approach outperformed human expert pilots and achieved a second-place rank in the ADT championship event. \end{abstract}

\begin{IEEEImpStatement}
Significant performance milestones in reinforcement learning have been achieved in recent years, with autonomous agents demonstrating super-human performance across a wide variety of tasks. Before these algorithms can be extensively deployed in real-world defense applications, a greater level of trust must first be achieved. ADT was an important step towards developing the trust necessary to operationalize these algorithms, by demonstrating their effectiveness on a foundational yet relevant problem in a high-fidelity simulation environment. Developed for the program, our hierarchical reinforcement learning agent was designed alongside of and competed against active fighter pilots, and ultimately defeated a graduate of the United States Air Force's F-16 Weapons Instructor Course in match play.
\end{IEEEImpStatement}


\newpage
\section{Introduction}

\IEEEPARstart{T}{he} Defense Advanced Research Projects Agency's (DARPA) Air Combat Evolution (ACE) program, seeks to build and enhance trust in 
autonomy for air-to-air combat and ultimately culminated in a live flight exercise with full-scale aircraft. The AlphaDogfight Trials (ADT) were created 
as a precursor to the larger ACE program in order to minimize risk. For ADT, teams proposed algorithmic solutions for one-vs-one within visual range (WVR) air combat, 
which is also known as a \emph{dogfight}. Solutions submitted to ADT ranged from rule-based systems to end-to-end machine learning, and eight finalists were selected 
to perform on the program. Each team's solution was tested in simulated dogfights in a high-fidelity simulation environment that leveraged an F-16 flight dynamics model (FDM). Scrimmages were held against a variety of opponents including DARPA-provided agents exhibiting various behaviors (e.g., cruise missile-like behavior), other finalist's agents, and expert human fighter pilots.

In deployment, autonomous flight control software such as autopilots and terrain avoidance systems are currently limited to deterministic and rule-based 
approaches, which can be attributed to a general lack of trust in AI-based autonomy. Before any pilot can advance to more complicated missions such as suppression of air defenses, escorting, and point protection, they must first master the basic flight maneuvers required for WVR air combat. For this reason, ACE selected the dogfight as the starting point for developing trust in autonomous air-combat systems.

In this article, we describe the simulation environment, the design and training of our agent, and the evaluation results showing the evolution of the agent ultimately submitted to the competition. Our approach uses hierarchical reinforcement learning (RL) that dynamically selects from a set of policies which are specialized for various aspects of an engagement. 
Our agent defeated a graduate of the United States Air Force (USAF) F-16 Weapons Instructor Course in match play with a perfect record (5 W - 0 L) and achieved 
$2^{nd}$ place overall in the final tournament.

The main contributions of this work are as follows:

\begin{itemize}
    \item Successful application of RL to train an agent capable of dogfighting in a high-fidelity simulation environment, at a level that exceeds human experts. In particular, we develop a hierarchical deep RL agent boosted by a variety of modern techniques including soft actor-critic, curriculum learning, distributed training, priority experience replay, and n-step returns.
    
    \item Detailed descriptions of reward shaping optimized for the dogfight scenario (Sections \ref{subsec:LL_policies}). 
        
    \item Proposal of a novel hierarchical mechanism using a high-level policy selector to optimally switch between previously trained low-level policies given the current context of the engagement (\ref{subsec:selector}). We provide a set of analyses demonstrating the effectiveness of this approach (\ref{subsec:effective_selector}).

    \item Presentation of training and evaluation frameworks along with strategies to develop a high-performance autonomous agent (Section \ref{sec:training}).
    
\end{itemize}

\section{Related Work}

Research on algorithms for autonomous air combat has been active since the 1950s\cite{isaacs}. Some approaches sought to use expert knowledge to formulate counter maneuvers as rule-based methods for a variety of positional contexts \cite{molineaux2010goal} \cite{burgin1988}. Other efforts have codified air-to-air scenarios as optimization problems to be solved computationally\cite{burgin1988, mcgrew2010, rodin1992, virtanen2006, austin1990}. 

Some research relies on building utility functions over a discrete set of actions and utilizing game theory~\cite{virtanen2006, austin1990}. Other 
approaches employ various forms of dynamic programming (DP) \cite{mcgrew2010, rodin1992, mcmanus}. In many of these approaches, trade-offs are made 
in the complexity of the algorithm and environment representation to reach approximately optimal solutions within a reasonable time 
frame \cite{virtanen2006, austin1990, mcgrew2010, rodin1992, mcmanus}. One notable work used Genetic Fuzzy Trees to develop an agent that 
defeated a USAF weapon school graduate in the USAF's Advanced Framework for Simulation, Integration and Modeling (AFSIM)~\cite{ernest}. 

More recently, deep RL has been applied to this problem~\cite{zhang2020, chen2020, Xu2019, yang2019, kong2020, wang2020improving, Pope2021, Yoo2021, Sun2021}. For example, Yang \textit{et al.} \cite{yang2019} leveraged deep-q networks to train an agent that selects from a collection of discrete maneuvers in a custom 3-D environment. Zhang \textit{et al.} \cite{zhang2020} evaluated a variety of RL algorithms applied to air missions modeled in AFSIM and found agents capable of learning cooperative strategies. Yoo \textit{et al.} leveraged recurrent neural networks to effectively predict opponent trajectories \cite{Yoo2021}. Sun \textit{et al.} developed a multi-agent RL algorithm with a hierarchical decision network that allows for the discrete selection of basic maneuvers (e.g., climb, turn, descend, etc.) and continuous actions to discipline those maneuvers \cite{Sun2021}. 

Most of the deep RL approaches surveyed either abstract the action space to high-level behaviors or rely on low-fidelity simulation environments. The former approach limits the agent's expressiveness and potentially its effectiveness, and the latter approach often bears little resemblance to real-world applications ~\cite{zhang2020, chen2020, Xu2019, yang2019, kong2020, wang2020improving}. The simulation environment used in this work, JSBSim, is uniquely high-fidelity in comparison to those used in related works. It provides an FDM of an F-16 aircraft with six degrees of freedom that accepts inputs to the flight control system. JSBSim is open-source software that is generally considered to be very accurate for modeling aerodynamics and the physics of flight \cite{vogeltanz2016, chen2009mobile}. In this article, we outline the design of an RL agent with hierarchical, continuous control of the aircraft that is capable of executing highly competitive tactics within this high-fidelity environment.

Independent of autonomous air combat research, there has been significant research on the topic of hierarchical RL \cite{dayanHinton1993, barto2003}. Dividing a complex task into smaller sub-tasks is at the core of many approaches ranging from classic divide-and-conquer algorithms to generating sub-goals in action planning \cite{schmidhuber1990}. In RL, temporal abstraction of state sequences converts the problem into a semi-Markov Decision Process (SMDP) \cite{sutton_etal_1999}. The approach is to abstract the agent's actions using macro-actions (routines), which are composed of primitive actions. This allows for modeling the agent at multiple levels of abstraction. This approach has a parallel with the hierarchical structure of human and animal learning \cite{botvinich2012}. The concept of hierarchical RL has generated important advancements such as option-learning \cite{comanici}, universal value functions \cite{schaul2015}, option-critic \cite{bacon2017}, FeUdal networks \cite{vezhnevets2017}, data-efficient hierarchical RL known as HIRO \cite{nachum} and many others. The main advantages of hierarchical RL are transfer learning (re-using previously learned skills and sub-tasks for new objectives), scalability (decomposition of large problems into smaller ones and/or avoiding the curse of dimensionality) and generalization (the combination of smaller sub-tasks to generate new skills and avoiding super-specialization) \cite{berliac2019}. 

In this article, we use a policy selector to select policies fitted to a current environmental sub-space. This approach resembles options learning algorithms \cite{comanici}, and is closely related to prior work on sub-policy selection methods~\cite{frans2018} in which sub-policies are hierarchically structured to perform new tasks. Sub-policies are primitives that are pre-trained in a similar environment but with different rewards. Our policy selector (similar to the master policy in \cite{frans2018}) learns to optimize a global reward given a set of \emph{low-level policies}. Low-level policies are specialized, pre-trained policies. However, unlike previous work which focused on meta-learning, our main objective is to learn to dog-fight optimally against different opponents, by dynamically switching between the low-level policies. Because of the complexity of the environment and the task, we do not iterate between the training of the policy selector and the low-level policies, i.e., the low-level policy agents' parameters (weights) are not updated in tandem with training the policy selector.

\section{Background}

\subsection{Reinforcement Learning}
\label{subsec:rl}

In short, a reinforcement learning (RL) agent is concerned about learning what actions to take to maximize a numerical reward signal given the current state of an environment\cite{sutton_barto2018}. Reward signals are provided by the environment and the learning is performed in an iterative way by trial-and-error. Therefore, reinforcement learning is different from both supervised and unsupervised learning. Supervised learning is performed on a training set of labeled examples provided by an external supervisor. Unsupervised learning does not work with labels but instead looks to find some hidden structure in the data \cite{mitchell1997}. 
One of the main challenges in reinforcement learning is to manage the trade-off between exploration and exploitation. As the RL agent interacts with the environment through actions, it starts to learn the choices that eventually return high reward. Naturally, to maximize the reward it receives, the agent should exploit what it has already learned by selecting the actions that resulted in high rewards. However, to discover the optimal actions, the agent has to take the risk and explore new actions that may lead to higher rewards than the current best-valued actions. 

\subsection{Maximum Entropy Reinforcement Learning}
\label{subsec:entropy}

In RL, knowing the best way to explore while exploiting is non-trivial, environment-dependent, and still an active area of research \cite{hong_etal_2018}. Maximum entropy RL theory provides a principled way to address this particular challenge, and has been a key element in many recent RL advancements, providing improved exploration and faster learning \cite{thomas2014,schulman2017a,haarnoja2017,haarnoja2018a,haarnoja2018b, ziebart2010}. 
Given a Markov decision process (MDP) with a set of states $S$, a set of actions $A$, a transition function $T$ and a reward function $R$, forming a tuple $<S,A,T,R>$ \cite{puterman1994}, a stochastic policy $\pi: S \rightarrow A$ is a mapping from states to probabilities of selecting each possible action, where $\pi(a|s)$ represents the probability of choosing action $a$ given state $s$. In maximum entropy RL, as in an MDP, the goal is to find the optimal policy $\pi^*$ that provides the highest expected sum of rewards, while additionally maximizing the entropy of each visited state, leading to the expression \cite{ziebart2010}: 
\begin{equation}
\pi^*=\argmax_{\pi} \sum_{t} \EX_{(s_t,a_t) \sim \rho_{\pi}}  [r_t+\alpha \mathcal{H}(\pi(\ \cdot \ |s_t))] ,
\label{eq:optimal-policy-max-H}
\end{equation}
where $\alpha$ is the temperature parameter that controls the stochasticity of the optimal policy, $\rho_{\pi}$ is the state-action marginal of the trajectory distribution induced by the policy, $r_t$ is a shorthand for the environmental reward  $r(s_t,a_t)$ at time $t$, and $\mathcal{H}(\pi)$ represents the entropy of the policy, $\EX[-log(\pi(a|s))]$. 
This approach allows a state-wise balance between exploitation and exploration. For states with high reward, a low entropy policy is permitted while, for states with low reward, high entropy policies are preferred, leading to greater exploration. The discount factor $\gamma$ is omitted in the equation for simplicity since it leads to a more complex expression for the maximum entropy case \cite{thomas2014}. But it is required for the convergence of infinite-horizon problems, and it is included in our final algorithm. Maximum entropy RL has conceptual and practical advantages. The entropy term leads the policy to explore more widely, and it also allows near-optimal behaviors. For instance, the agent will prefer two equally attractive actions (higher entropy) than a single action that restricts exploration. In practice, improved exploration and faster learning have been reported in many prior works \cite{schulman2017a}, \cite{haarnoja2017}.
\subsection{Soft Actor-Critic}
\label{subsec:ac}
Soft actor-critic (SAC) \cite{haarnoja2018a} is one of the most successful maximum entropy RL methods and has become a common baseline algorithm in most popular RL libraries, outperforming state-of-the-art methods \cite{haarnoja2018a,haarnoja2018b}. Like the deep deterministic policy gradient (DDPG) approach \cite{lillicrap2016}, SAC is a model-free and off-policy method, using a replay buffer, where the policy and value functions are approximated using neural networks. In addition, it incorporates a policy entropy term into the objective function facilitating exploration, similar to soft Q-learning \cite{haarnoja2017}. Similar to trust region policy optimization (TRPO) \cite{schulman2015} and proximal policy optimization (PPO) \cite{schulman2017b}, SAC uses a stochastic policy and is known to be more stable than DDPG. In short, SAC combines the best of DDPG (sample efficiency) and TRPO/PPO (stability through stochastic policies). As expressed in Equation \ref{eq:optimal-policy-max-H}, the SAC policy/actor is trained with the objective of maximizing the expected cumulative reward and the action entropy at a particular state. The critic is the soft Q-function and, following the Bellman equation, is expressed by: $Q(s_t, a_t) = r_t +\gamma \EX_{s_{t+1} \sim p} [V(s_{t+1})]$,
where $p$ represents the state transition probability, and the soft value function is parameterized by the Q-function:
$V(s_{t+1}) =  \EX_{a_t \sim \pi} [Q(s_{t+1}, a_{t+1}) - \alpha \log \pi(a_{t+1} \vert s_{t+1})]$.
The soft Q-function is trained to minimize the following objective function given by the mean squared error between predicted and observed state-action values:
\begin{equation}
J_Q = \EX_{(s_t,a_t) \sim \mathcal{D}} \left[\tfrac{1}{2}\big( Q(s_t, a_t)-\hat{Q}(s_t, a_t) \big)^2 \right]
\label{eq:objective-Q}
\end{equation}
where 
\begin{equation}
\hat{Q}(s_t, a_t) = r_t + \gamma \EX_{s_{t+1} \sim p}[\bar{V}(s_{t+1})],
\label{eq:objective-Q}
\end{equation}
$\mathcal{D}$ denotes the replay buffer, and $\bar{V}$ is the target value function \cite{mnih2015}.
Finally, the policy is updated to minimize the KL-divergence between the policy and the exponentiated state-action value function \cite{haarnoja2018b}, and can be expressed by:
\begin{equation}
J_{\pi} = \EX_{s_t \sim \mathcal{D}} \left[ \EX_{a_t\sim \pi} [\alpha \log \pi(a_t \vert s_t) - Q(s_t, a_t)] \right].
\label{eq:objective-pi}
\end{equation}
\subsection{Reward Shaping}
\label{subsec:shaping}
Similar to many real-world problems, the reward provided by the ADT simulation environment for the dogfighting task is sparse. Success depends on achieving a narrow pointing angle and close proximity to the opponent. No other reward such as flying smoothly or getting close to the opponent is natively provided by the environment. This poses a significant challenge for training the RL agent because it may never receive a reward signal, and hence never learn how to perform the task. Therefore, reward shaping~\cite{Mataric1994, Ng1999} is necessary to provide more feedback during training and exploit implicit domain knowledge to accelerate learning. In spite of recent advances to overcome reward sparsity such as replay buffers~\cite{lillicrap2016}, universal value function approximators for setting intermediate subgoals~\cite{schaul2015} and hindsight experience replay for learning from failures~\cite{Andrychowicz2015}, it is still difficult to successfully achieve sample efficiency. Hence, learning from sparse rewards continues to be an active area of research~\cite{Rauber2019, Agarwal2019}.

In this work, given the computationally intensive nature of the environment (demanding sample efficiency) and the complexity of the task, we performed reward shaping informed by domain knowledge in order to compose dense reward signals. This choice provided many advantages: sample efficiency, faster learning and the ability to encode expert knowledge and define low-level agents with specific desired behaviors. In the subsequent sections, we describe each of the shaped rewards in detail.

\section{ADT Simulation Environment}
\label{sec:adt}

The simulation environment used to model the dogfighting scenario was developed by the Johns Hopkins University Applied Physics Lab (JHU-APL).
It is an extension of an open-source RL gym environment\footnote{https://github.com/Gor-Ren/gym-jsbsim}, which was designed for single aircraft 
flight and leverages JSBSim, a high-fidelity open-source FDM \cite{jsbsim}. JHU-APL extended this work to accommodate multiple 
F16 aircraft and integrated a more robust visualization engine. 

The observation space provided by the environment includes ownship position (local plane coordinates, velocity, and acceleration), attitude (Euler angles, rates, and accelerations), aerodynamics (alpha and beta angles), and other miscellaneous values (fuel load, thrust, control surface deflection, and health). Additionally, the position (local plane coordinates and velocity), and attitude (Euler angles and rates), and health of the opponent are given as well. All state information from the simulation environment is provided without modeled sensor noise. 

We experimented with a variety of observation spaces throughout the course of development, including the removal or normalization of default simulation state values as well as the addition of derived states. Derived states included egocentric and allocentric polar coordinates as well as airspeed. Additionally, we also experimented with including state changes (deltas) in the observation spaces of the agents with varying degrees of time between observed state values.

A complete list of the observation space of the simulation as well as the observation space of the agents described in this article can be found in the supplemental material.

Interactions with the environment occur at a rate of 50 times per simulation second. The action space is composed of the main inputs to the F-16's flight control system (aileron, elevator, rudder, and throttle) and is continuous. An agent inflicts damage and reduces the health of its opponent whenever the opponent is within the agent's Weapons Engagement Zone (WEZ).

\begin{figure}[h!]
\centering
\includegraphics[width=0.8\columnwidth]{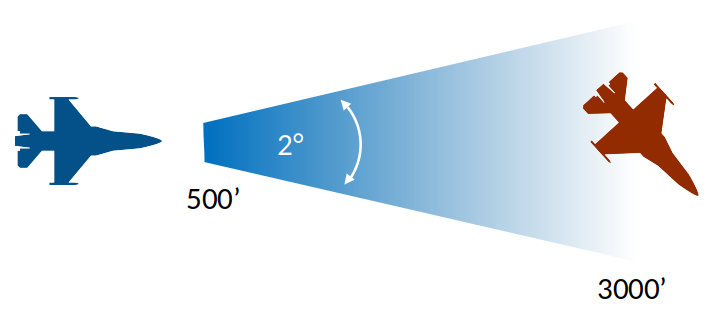}
\caption{Weapon Engagement Zone (WEZ).}
\label{fig2}
\end{figure}

The WEZ is defined as the locus of points in between 500-3000 ft that also lie within a spherical cone of 2\textdegree{} aperture that 
extends out of the plane's nose (Figure \ref{fig2}). Though the agent is not truly taking shots on its opponent, throughout this article we 
use the term ``gun snap'' to refer to attaining this geometry.

The damage per second incurred from within the WEZ is given by
\begin{equation}
D_{wez} = \begin{cases}
      \frac{3000-d}{2500} & 500 ft \leq d \leq 3000 ft \\
      0 & otherwise \\
\end{cases}\label{eq:dwez}
\end{equation}
\noindent where $d$ is the distance between the two aircraft.

The reward given by the environment is equal to the average amount of total damage an agent has dealt to its opponent and is given by
\begin{equation}
r_t = \mathds{E}_{t'\in[0,T]}[\sum_{n=0}^{t}D_{wez}(t)]\\
\end{equation}

\noindent where $r_t$ is the reward at time step $t$ and $T = 300$, the maximum duration of an engagement.
The engagement ends once the simulation time reaches 300 seconds or if either aircraft reaches zero health. An agent ``wins'' when its opponent's health reaches zero. This can happen if sufficient damage is incurred via gun snaps or the minimum altitude (hard deck) of 1000 feet is breached. The engagement is considered a draw if the simulation times out.
\section{Methods}

\subsection{Agent Architecture}
\label{sec:architecture}

\begin{figure}[h!]
\centering
\includegraphics[width=0.95\columnwidth]{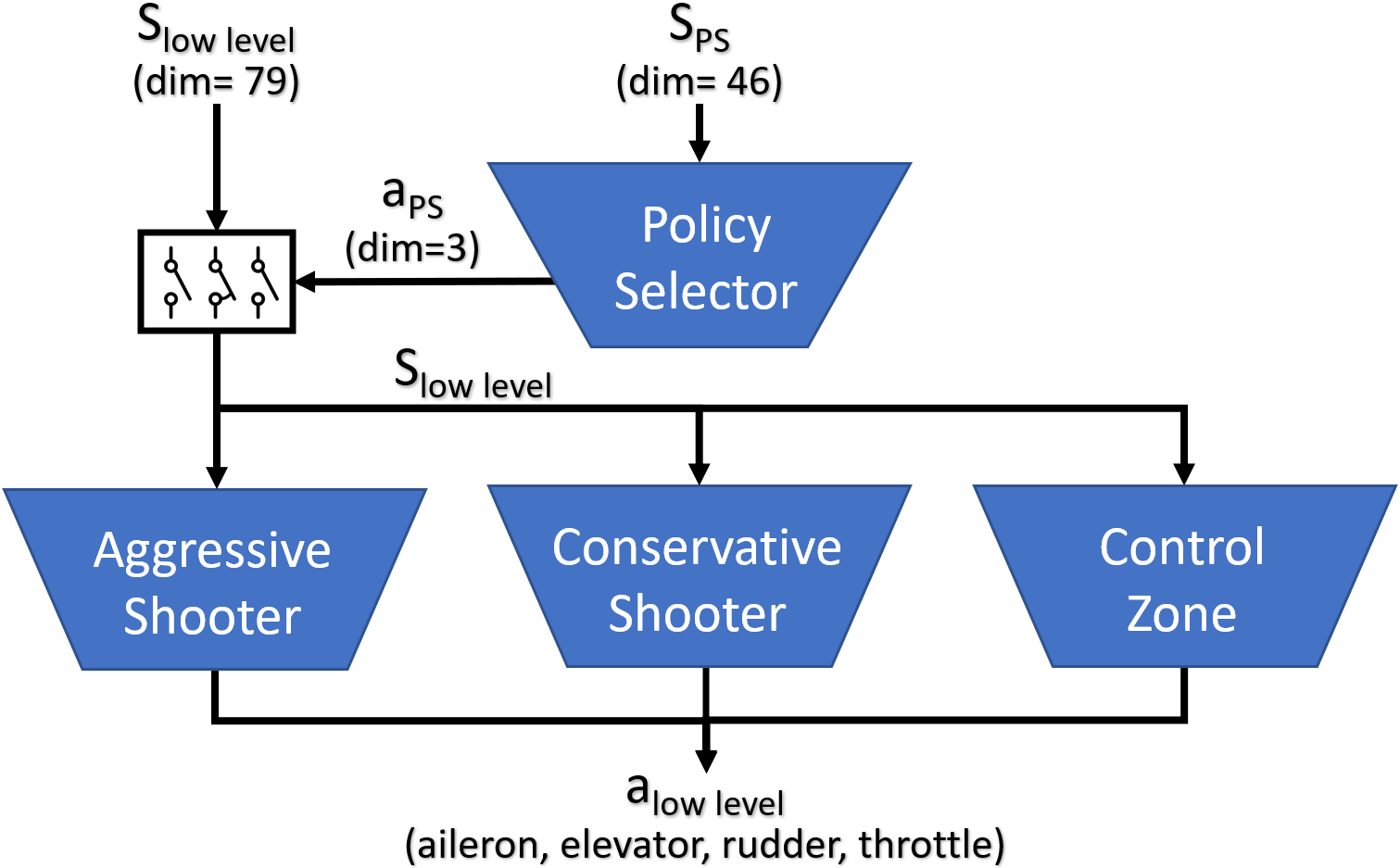}
\caption{High-level architecture of PHANG-MAN agent.}
\label{fig4}
\end{figure}

Our agent, PHANG-MAN (Policy Hierarchy for Adaptive Novel Generation of MANeuvers), is composed of a 2-layer hierarchy of policies. There is an array of policies at the lower level of the hierarchy that have each been trained to excel in a particular region of the state space. A singular policy at the higher level of the hierarchy selects which low-level policy to activate given the current context of the engagement, as shown in Figure \ref{fig4}. We refer to this policy as the policy selector. This architecture allows for the high-level and low-level rewards to be decoupled, and it provides the ability to extend existing hierarchical agents by incorporating additional low-level policies that address specific skill gaps. If the behavior of a low-level policy aids in achieving the policy selector's goal, the policy selector will learn which regions of the state space to employ it.

\subsection{Low-Level Policies}
\label{subsec:LL_policies}
The low-level policies interact with the environment at a frequency of 50 Hz and are trained via Soft Actor-Critic (SAC) \cite{haarnoja}. The same observation space is used for all three policies, and a detailed list of the quantities is provided in the supplemental material. Multilayer perceptrons (MLP) with a single hidden layer of 12288 neurons are used for all policies and Q-functions. Wide and shallow MLP networks such as these have been shown to perform better than narrower and deeper MLP networks containing the same number of neurons \cite{largernetworks}. What distinguishes the low-level policies are the unique reward functions and the range of initial conditions used during their training. The reward functions are the sum of various independent components that each promote a target behavior.\\
\begin{figure}
\centering
\includegraphics[width=0.99\columnwidth]{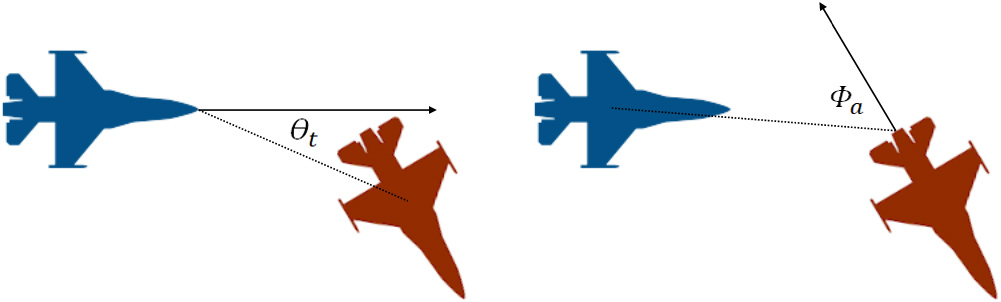}
\caption{Depiction of track angle, $\theta_{t}$, (left) and adverse angle, $\phi_{a}$ (right).}
\label{fig:geometry}
\end{figure}
To facilitate the description of the reward function components, we introduce the following:
\begin{itemize}
\item Track angle, $\theta_{t}\in[0,180]$, is the angle in degrees between the agent's aircraft nose and the center of the opponent's aircraft, shown in Figure \ref{fig:geometry} (left). We will use the normalized value $\bar{\theta}_{t} = \tfrac{\theta_{t}}{180}$.
\item Adverse angle, $\phi_{a}\in[0,180]$ is angle in degrees between the opponent's tail and the center of the agent's aircraft, shown in Figure \ref{fig:geometry} (right). We will use the normalized value $\bar{\phi}_{a} = \tfrac{\phi_{a}}{180}$.
\item The distance between the two aircraft, $d$, measured in feet.
\item Closure rate, $v_{c}$, is the rate of change of $d$ and is measured in feet per second.
\item  The logistic function,
\begin{equation}
    S(x, \alpha, x_{0})=\frac{1}{1+e^{-\alpha (x-x_{0})}},
\end{equation}
parameterized by growth rate, $\alpha$, and midpoint, $x_{0}$.
\end{itemize}
\begin{table}
\centering
\caption{Reward function component equations}
\label{table:rewards}
\small
\hrule
\vspace{6px}
\begin{align}
r_{\phi_{a}} &= -\bar{\phi}_{a} \label{eq:radverse}\\[6px]
r_{\theta_{t}} &= -\bar{\theta}_{t}^{\lambda} \label{eq:rtrack}\\[6px]
r_{\text{$rel.$ $pos.$}} &= (\bar{\theta}_{t}-2) \cdot S(\bar{\phi}_{a}, 18, \tfrac{1}{2})-\bar{\theta}_{t} + 1 \label{eq:relative_position}\\[6px]
r_{closure} &= \frac{v_{c}}{500} \cdot [1-S(\bar{\phi}_{a}, 18, \tfrac{1}{2})] \cdot S(d, \tfrac{1}{500}, 2900) \label{eq:closure}\\[6px]
r_{gunsnap(blue)} &= \Gamma_{B}(d) \cdot [1-S(\bar{\theta}_{t}, 1e5, \tfrac{1}{180})] \label{eq:gunsnap_blue}\\[6px]
r_{gunsnap(red)} &= -\Gamma_{R}(d) \cdot S(\bar{\phi}_{a}, 800, \tfrac{178}{180}) \label{eq:gunsnap_red}\\[6px]
r_{deck} &= -4 \cdot [1-S(h, \tfrac{1}{20}, 1300)] \label{eq:deck}\\[6px]
r_{\text{$too$ $close$}} &= -2 \cdot [1-S(\bar{\phi}_{a}, 18, \tfrac{1}{2})] \cdot S(d, \tfrac{1}{50}, 800) \label{eq:too_close}
\end{align}
\vspace{2px}
\hrule
\end{table}
The equations of each of the reward function components are provided in Table \ref{table:rewards}, and qualitative descriptions are as follows:
\begin{itemize}

\item $r_{\phi_{a}}$ penalizes the agent for having a non-zero adverse angle.

\item $r_{\theta_{t}}$ penalizes the agent for having a non-zero track angle. In Equation \ref{eq:rtrack}, $\lambda$ is unique across policies.

\item $r_{\text{$relative$ $position$}}$ rewards the agent for attaining the ideal dogfighting geometry in which its track angle and adverse angle are minimized (i.e., pointing directly at its opponent from directly behind its opponent's tail). It also penalizes the agent for allowing its opponent to do the same.

\item $r_{closure}$ rewards the agent for getting closer to its opponent from behind and penalizes the agent for allowing its opponent to get closer from behind. It is scaled by distance, in order not to encourage overshooting.

\item $r_{gunsnap(blue)}$ rewards the agent when it achieves a minimum track angle and is within a particular range of distances. In Equation \ref{eq:gunsnap_blue}, $\Gamma_{B}(d)$ is the range factor that determines the distances at which blue gun snap rewards have the greatest magnitude. It is given by
\begin{equation}
\Gamma_{B}(d) =
    \begin{cases} 
        \beta_{B}(d) \cdot S(d, \tfrac{1}{50}, 1000)  & d < 1950 ft \\
        \beta_{B}(d) \cdot [1-S(d, \tfrac{1}{50}, 2900)] & d \geq 1950 ft \\       
    \end{cases}\\
\end{equation}
where $\beta_{B}(d)$ is the aggressiveness factor for blue gun snaps. $\beta_{B}(d)$ is unique across policies and defined later for each instance.

\item $r_{gunsnap(red)}$ penalizes the agent when its opponent achieves a minimum track angle and is within a particular range of distances. In Equation \ref{eq:gunsnap_red}, $\Gamma_{R}(d)$ is the range factor for red gun snaps and is given by
\begin{equation}
\Gamma_{R}(d) =
    \begin{cases} 
        \beta_{R}(d) \cdot S(d, \tfrac{1}{35}, 400) & d < 2250 ft \\
        \beta_{R}(d) \cdot [1-S(d, \tfrac{1}{200}, 4100)] & d \geq 2250 ft \\       
    \end{cases}\\
\end{equation}
$\beta_{R}(d)$ is the aggressiveness factor for red gun snaps, which is unique across policies and defined later for each instance.

\item $r_{deck}$ penalizes the agent for violating a minimum altitude threshold. It is defined by where $h$ is the height above the ground in feet.

\item $r_{\text{ $too$ $close$}}$ penalizes the agent for violating a minimum distance threshold when below a maximum adverse angle and is meant to discourage overshooting when pursuing.
\end{itemize}

\begin{figure*}[h!]
\centering
\includegraphics[width=2.04\columnwidth]{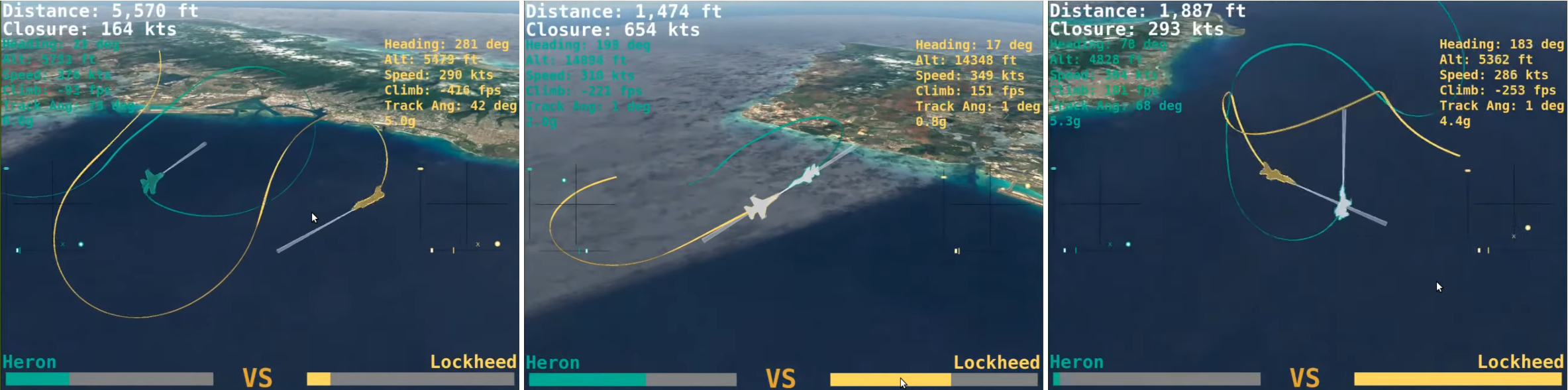} 
\caption{Typical behavior of the low-level policies driven by reward shaping. Control Zone (left), Aggressive Shooter (middle), Conservative Shooter (right).}
\label{fig:typical_LL_behavior}
\end{figure*}

\vspace{4px}
\noindent A description of each of the low-level policies follows:

\vspace{8px}
\textit{Control Zone (CZ)}: The CZ policy attempts to attain a geometry in which there is nothing the opponent can do to deny the agent's positional 
advantage. This position is known as the ``control zone'' in the fighter pilot community and is roughly defined as the cone-like area projecting out of the tail of an aircraft, 20 degrees wide at 2000 feet and 40 degrees wide at 4000 feet \cite{navybfm}. Insights from fighter pilots helped to design the CZ policy's multi-dimensional reward function, which is dependent on track angle, adverse angle, distance to opponent, height above the hard deck, and closure rate. The CZ reward function is equal to the sum of the rewards given by Equations \eqref{eq:relative_position}-\eqref{eq:too_close}, with
$\beta_{B}(d)=3$ and $\beta_{R}(d)=-3$.

\begin{figure}[h!]
\centering
\includegraphics[width=0.68\columnwidth]{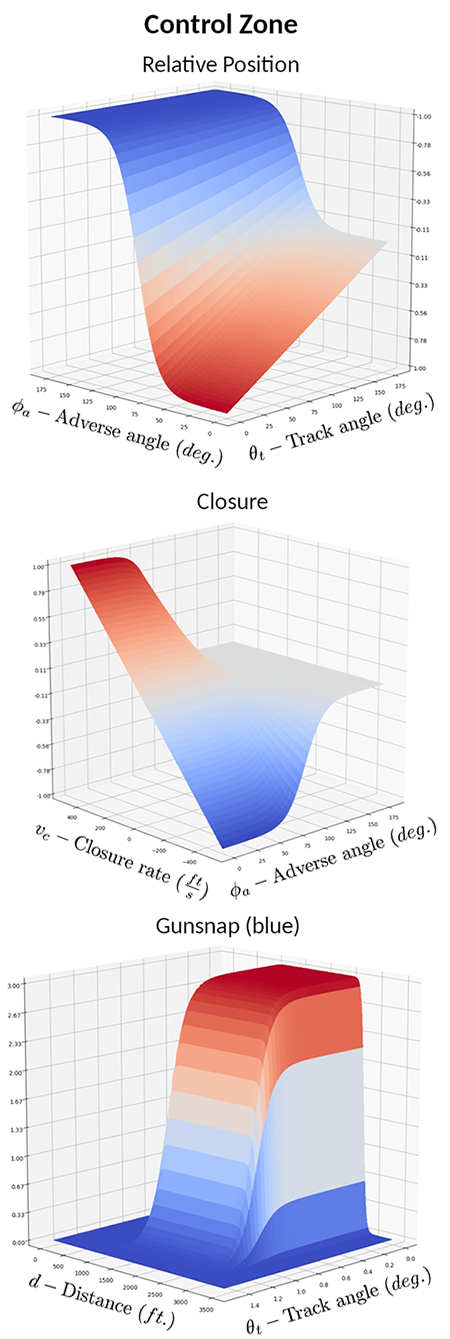}
\caption{$r_{\text{ $rel.$ $pos.$}}$, $r_{closure}$, and $r_{gunsnap(blue)}$ components of the CZ policy's reward function.}
\label{fig:CZ}
\end{figure}

The surfaces of select reward function components of the CZ policy are shown in Figure \ref{fig:CZ}. The CZ policy was trained with the widest range of initial conditions of all the low-level policies and included uniformly random positions, Euler angles, and body-frame velocities. Representative behavior of the CZ policy is show in Figure \ref{fig:typical_LL_behavior} (left).

\textit{Aggressive shooter (AS) and Conservative shooter (CS)}: The most impactful difference between the two shooter policies and the CZ policy is that
$r_{\text{ $rel.$ $pos.$}}$ is replaced with $r_{\theta_{t}}$ \eqref{eq:rtrack}, with $\lambda=\tfrac{4}{9}$.
Since there is no dependence on $\phi_{a}$, the shooter policies learn to attack from all sides, including head on.

The differences between the two shooter policies are relatively subtle. The most notable difference between the two is the AS policy gun snap reward 
is greater in magnitude at closer distances whereas the CS policies gun snap reward is constant regardless of distance. Qualitatively as a result, the AS 
policy takes shots that will yield the most damage but potentially leave it susceptible to counter maneuvers, and the CS policy prioritizes maintaining 
an offensive scoring position, even though the magnitude of the points scored may be low. There is a similar distinction on the defensive side as well. 
The CS policy avoids gun snaps equally from all distances, making it a relatively aggressive evader, whereas the AS policy avoids gun snaps from close range 
more than those from farther away, making it a less aggressive evader.

\begin{figure}[h!]
\centering
\includegraphics[width=0.67\columnwidth]{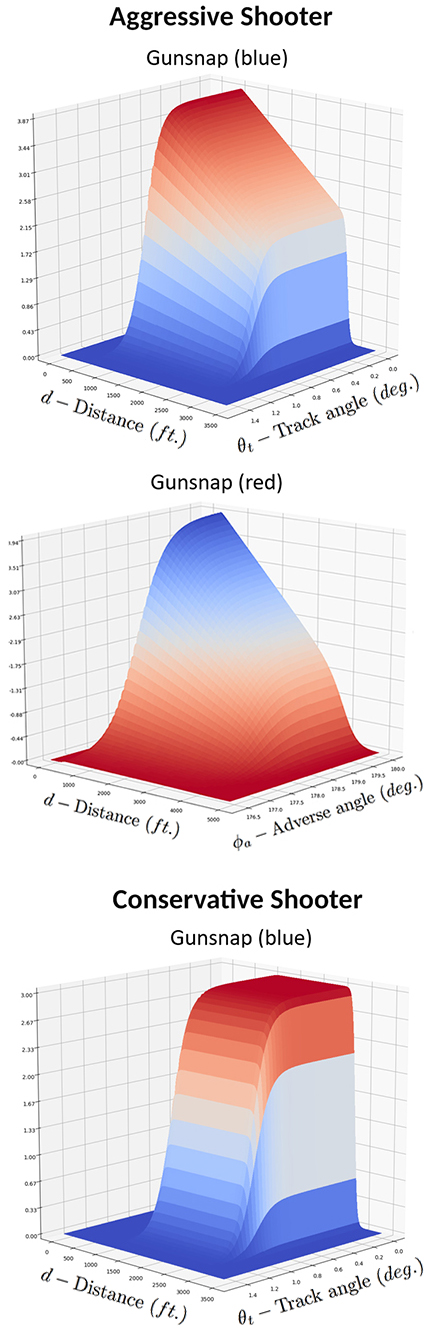}
\caption{$r_{gunsnap(blue)}$ and $r_{gunsnap(red)}$ components of the AS policy's reward function (top) and $r_{gunsnap(blue)}$ component of the CS policy's reward function (bottom).}
\label{fig:CS}
\end{figure}

Both the AS and CS policy were trained with initial conditions that place the opponent in the agent's WEZ or vice versa. This increased the ratio of WEZ to non-WEZ memories stored in the replay buffer, biasing the agent's learning toward regions of the state space where offensive and defensive gun snap rewards are the most significant. Additionally, since the time from initial gun snap to episode conclusion was generally observed to be short (sometimes only 1 step), the total health of both aircraft was increased by 10x, to further increase in the ratio of WEZ to non-WEZ memories stored in the replay buffer. The reward function of both the AS and CS are given by
\begin{equation}
    r_{AS/CS} = r_{\theta_{t}} + r_{gunsnap(blue)} + r_{gunsnap(red)} + r_{deck}.\label{eq:asps}
\end{equation}
Although $r_{AS/CS}$ is the sum of the same components for both the AS and CS policy, the reward components $r_{gunsnap(blue)}$ and $r_{gunsnap(red)}$ themselves are unique. Specifically, for the AS policy,
\begin{equation}
    \beta_{B}(d) =
    \begin{cases} 
        2(\tfrac{2900 - d}{1900} + 1)  &  1000 ft \leq d \leq 2900 ft \\
        2 & otherwise \\       
    \end{cases}\\
    \label{eq:gsslopeblue}
\end{equation}
and
\begin{equation}
\beta_{R}(d) =
    \begin{cases} 
        -2(\tfrac{4100 - d}{3700} + 1)  &  400 ft \leq d \leq 4100 ft \\
        -2 & otherwise \\         
    \end{cases}.\\
    \label{eq:gsslopered}
\end{equation}
For the CS policy $\beta_{B}(d)=3$ and $\beta_{R}(d)=-3$. The surfaces of select reward function components of the AS and CS policies are shown in Figure \ref{fig:CS}. Representative behaviors of the AS and CS policy are shown in Figure \ref{fig:typical_LL_behavior} (middle, right).\\

\subsection{Policy Selector (PS)}
\label{subsec:selector}

At the top of the hierarchy the policy-selector acts like an intelligent de-multiplexer, identifying which of the low-level policies to activate given the current context of the engagement. Our approach is similar to options learning \cite{comanici}, however we do not estimate the terminal conditions directly. Instead, the policy selector is evaluated periodically at a frequency of 10 Hz, which is every fifth action that the low-level policies make. This is much like the Meta Learning Shared Hierarchies architecture \cite{frans2018}. Unlike that approach, we independently train the low-level policies to have target behaviors and then freeze their parameters (weights), which allows training the policy selector without the complications associated with non-stationary policies \cite{nachum}. This simplifies the learning problem and permits training and re-use of agents in a modular fashion.

The policy selector, which has a discrete action space, is implemented using the same SAC algorithm as the low-level policies. Although an SAC implementation for discrete actions is viable for choosing low-level policies \cite{christodoulou}, the continuous version of SAC combined with argmax demonstrated better performance. This is unlikely to hold true for larger action spaces, where discrete action space sampling methods such as Gumbel-Softmax are critical to ensure proper exploration \cite{jang2016categorical}. The policy selector's state space is a subset of that used to train the low-level policies, where accelerations, control surface positions, and other quantities are omitted. The full list of values is given in the supplemental material.

The policy selector's reward function is given by,
\begin{equation}
    r_{PS} = r_{\theta_{t}} + r_{gunsnap(blue)} + r_{gunsnap(red)} + r_{deck}, \label{eq:r_ps}
\end{equation}
with $\beta_{B}(d)=5$, $\beta_{R}(d)=-5$, and $\lambda=1$. The dominant terms in $r_{PS}$, $r_{gunsnap(blue)}$ and $r_{gunsnap(red)}$, resemble the damage calculation 
within the environment \eqref{eq:dwez}, which is sparse. This ensures that the PS learns strategies that are well aligned with the scoring metric of the environment.
To facilitate learning and avoid the challenges associated with sparse rewards, we include a relatively smaller penalty that is directly proportional to track angle.
We also include $r_{deck}$ to prevent the PS from learning risky switching combinations that can cause the plane to fly too low, even though each of the low-level
policies have been trained to avoid the deck individually.

\begin{figure}[h!]
\centering
\includegraphics[width=1\columnwidth]{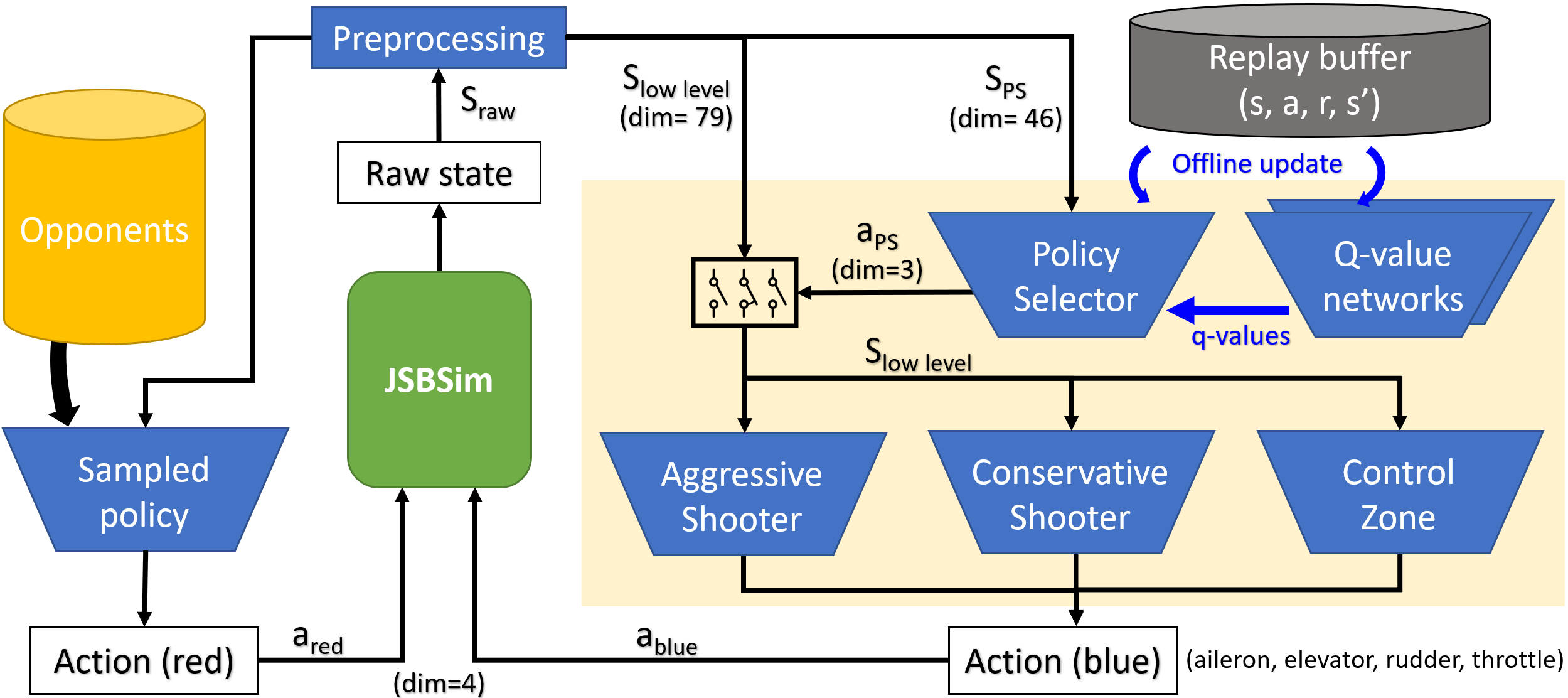} 
\caption{Training architecture of the PHANG-MAN agent.}
\label{fig:architecture}
\end{figure}

In Figure \ref{fig:architecture}, we summarize the hierarchical structure of the agent along with its training framework. Notice that a single-instance version of the offline updating is shown here.
The hyperparameters for the low-level policies and policy selector are given in Table \ref{table:hyper}.

\setcellgapes{2pt}
\newcolumntype{P}[1]{>{\centering\arraybackslash}p{#1}}
\begin{table}[htbp!]
\caption{List of common hyper-parameters used in the SAC training.}
\label{table:hyper}
\small
\makegapedcells
\begin{tabular}{p{0.28\textwidth}p{0.16\textwidth}}
\hline
\textbf{Parameter} & \textbf{Value} \\
\hline
Optimizer &	 Adam \\
Replay buffer size &	 5.0e6 \\
Number of hidden layers & 1 \\
Number of hidden neurons & 7168 (Policy Selector) \\ & 12288 (CZ, AS, CS) \\
Batch size & 256 \\
Learning rate & 2.0e-4 \\
Discount factor $\gamma$ & 0.99 \\
Soft $\tau$ & 1.0e-3 \\
Entropy target $\tau$ & -3 (Policy Selector) \\ & -4 (CZ, AS, CS) \\
Activation function & ReLU \\
Target update interval & 1 \\
\hline
\end{tabular}
\end{table}

\section{Training Framework and Evaluation}
\label{sec:training}

In this section, we describe the training and evaluation frameworks as well as additional training strategies used to develop the PHANG-MAN agent. 

\subsection{Training Framework and Hardware}

For this effort, a custom off-policy distributed training framework was developed in Pytorch (v1.3.1). The framework is similar to the Ape-X architecture \cite{Horgan2018} but enhanced to accommodate SAC. It is composed of distributed actors, a central learner, and a central prioritized experience replay (PER) buffer. A high-level architecture diagram can be found in supplemental material. Each actor has its own instance of the environment and a copy of the most recent policy weights. The centralized PER buffer managed the most significant experiences, while the central learner continuously updated the policy and value networks. A single training instance on an Amazon EC2 P3.16xlarge (64 CPUs, 8 V100 GPUs, 488 GiB RAM) instance could run 21 actors and was CPU limited due to the computationally demanding JSBSim environment. Training a policy selector network took up to a week, and the low-level policies took up to a month to achieve maximum performance.

\subsection{Training Strategy}
\label{subsec:training_strategy}
In addition to the diverse reward shaping and the distributed training framework, the following components were also critical to train agents and optimize 
them for the ADT competition:

\begin{itemize}

\item{Dynamic selection of opponents (curriculum learning): Agents initially trained only against opponents with scripted behaviors that were considered ``easy''. Once the agent achieved a win/loss ratio greater than 50\%, additional ``intelligent'' opponents were introduced, following a game-theoretic approach \cite{Tuyls2005,Garnelo2021}. These additional opponents included previous iterations of PHANG-MAN and individual low-level policies. Some of these agents had more defensive or offensive tactics, while others were meant to mimic the behavior of other competitors' agents. After the agent played every opponent 100 times and had an overall win/loss ratio greater than 50\%, opponents were sampled from a probability distribution proportional to the win/loss ratio of the last 100 matchups against every opponent. The distribution  was dynamically updated after each matchup. This created more matches against challenging opponents and prevented redundant matches against opponents that the agent was able to consistently defeat. The probability of playing any opponent was clipped to a minimum of $0.2\%$ and a maximum of $11.7\%$. Thirty different opponents were employed during training in addition to self-play.}

\item{Truncation of state and action values: Significantly different outcomes from identical scenarios were observed when deploying on hardware with a different GPU architecture than the training hardware (e.g. Volta vs. Tesla). Sometimes, these differences were large enough to cause a loss that would have otherwise been a win and vice versa. We observed identical inputs to identical networks that were running within identical docker containers producing different outputs, on the order of 1e-6. We speculate that the cause is due to different hardware-based floating point rounding algorithms across architectures, but further investigation is required. Regardless, when compounding these differences 50 times per simulated second, the F-16 FDM quickly drifts from its rightful state. To synchronize the performance of the agent when deployed on different GPU architectures, we trained all policies with the state values truncated to six decimals and action values truncated to two decimals.}

\item{Initial conditions: Some of the state variables such as relative position, altitude and velocity defined the positional advantage or disadvantage at the 
beginning of each episode. These initial conditions were classified into categories going from highly defensive to highly offensive positions and were selected 
according to the policy's specific goals. For example, the AS and CS shooter policies were mostly trained in highly offensive and defensive initial conditions 
whereas the CZ policy was trained across a more uniform range of initial conditions that balanced mixture included offensive, defensive and neutral.}

\item{Stochastic multi-step returns: For temporal distance learning, multi-step returns often lead to faster learning when a suitably tuned number of steps in 
the future is used \cite{Sutton1998}. Instead of tuning a fixed value, we defined the maximum number of steps in the future and uniformly sampled values up to 
that maximum, storing this value with every experience in the PER buffer. We found 25 steps to be an effective maximum for the low-level policies and 5 for 
policy selector, which accounts for 0.5 seconds of simulation time in both cases.}

\end{itemize}

\subsection{Performance Metrics during Training and Validation}
\label{subsec:performance_metrics}
\begin{figure}[h!]
\centering
\hspace*{-0.25cm}\includegraphics[width=0.9\columnwidth]{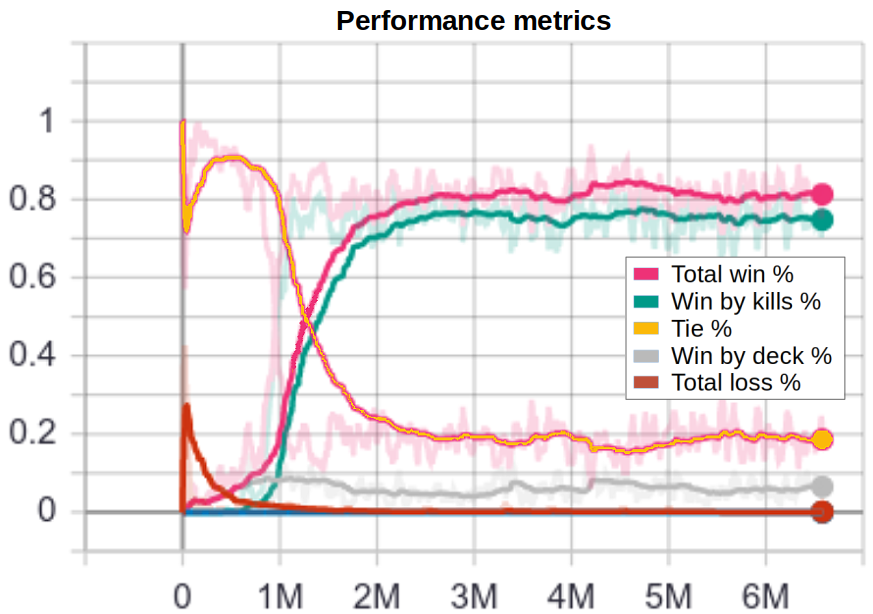} 
\caption{Example of performance metrics during training against the agent Bud-FSM.}
\label{fig_training_metrics}
\end{figure}

To closely monitor the agent's learning across multiple opponents throughout training, we computed performance metrics for each individual opponent. In Figure \ref{fig_training_metrics} we illustrate the performance against one particular opponent, a finite state machine named Bud-FSM. This graph is the result of an individual training run consisting of tens of thousands of engagements against a wide variety of opponents, in addition to Bud-FSM. Having a diverse set of opponents and the respective performance metrics was important to identify the strengths and weaknesses of the trained agents. These metrics also provided insight on how to effectively group complementary low-level policies together, into what we will refer to as a team. For example, a particular opponent that forced a substantial amount of deck hits indicated the need to include a team-member with better low-altitude tactics.

During the ADT events, competitor's agents were first evaluated according to their performance against the scripted agents developed by JHU-APL. Therefore, 
to select the best performing agents, we created a benchmark composed of five opponents and played twenty-five matches against each, with a fixed set of 
initial conditions ranging from defensive to offensive. These initial conditions closely resemble those from previous ADT events and are referred to as ADT initial conditions. The insights gained from these evaluations informed future reward function iterations.

\section{Experimental Results}
\label{sec:exp_results}

\subsection{Trained Low-Level Policies}

The training of low-level policies was monitored with the same performance metrics (Section \ref{subsec:performance_metrics}) used to evaluate the hierarchical RL agents. The goal was to train the best performing low-level policy within each category of behavior (CZ, AS, CS) shaped by the reward functions. 
Their behaviors were qualitatively evaluated by experts (former pilot, Air Force officer) watching a few playbacks, and reward functions were interactively shaped until achieving their final forms (Section \ref{subsec:LL_policies}). We tested all the low-level policies against scripted ADT agents and previous generations of the PHANG-MAN agent, using a variety of initial conditions. We observed that their performance was sensitive to initial positions and the particular opponent, implying that a Pareto type of dominance was not observed. For example, in head-to-head offensive initial positions, as expected, the AS policy performed better; while in neutral positions, in which positioning was important, the CZ policy performed better. 

In Table \ref{table:low-level}, we list all the trained low-level policies. Notice that for training CZ policies, we used $50\%$ random and $50\%$ offensive initial positions, while for shooter policies, we primarily focused on offensive positions. The rationale was to optimize training by narrowing down the possible state-space. Further, in Table \ref{table:evolution}, we show the performance of two low-level policies against several hierarchical policies. The most aggressive (highest number of wins by gun snap) and most defensive (lowest number of Losses) opponents are agents $\{7, 9b\}$ and $\{9b, 11a, 11b\}$, respectively. Interestingly, the CS policy performed better against aggressive than defensive agents with average 28 and 22 wins, respectively; while the AS policy performed slightly better against defensive than offensive agents, with average 22 and 19 wins, respectively.

\subsection{Effectiveness of the Policy Selector}
\label{subsec:effective_selector}
\begin{figure}[h!]
\centering
\hspace*{-0.25cm}\includegraphics[width=0.9\columnwidth]{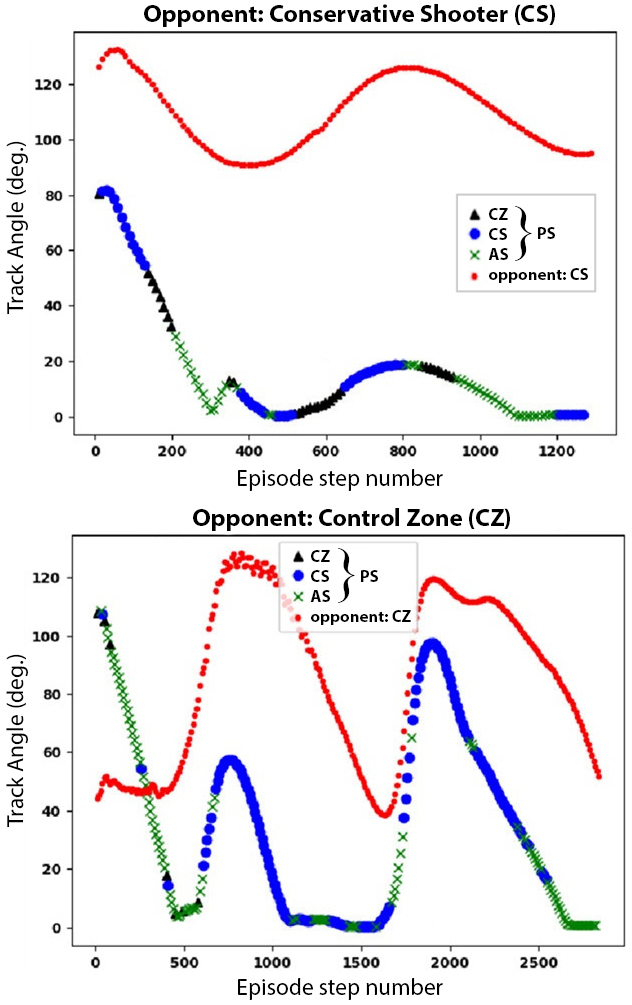}
\caption{Low-level policy selection throughout an episode. Policy selector (PS) with a team of 3 low-level policies: CZ, CS, and AS. Opponents: CS (top) and CZ (bottom). Each dot represents a selection made by the PS every 5 steps.}
\label{fig:LL_selection}
\end{figure}

In Figure \ref{fig:LL_selection}, examples of the policy selector's dynamic behavior are shown as a function of track angle, one of the most telling state variables. In the match against the CS policy (top), the agent exploits the initial positional advantage (smaller track angle), closes the distance while reducing the track angle, it overshoots once (step 500) but gets persistent gun snaps (steps 1100 to 1300) and ends the engagement victoriously.

\begin{figure}[h!]
\centering
\hspace*{-0.25cm}\includegraphics[width=0.9\columnwidth]{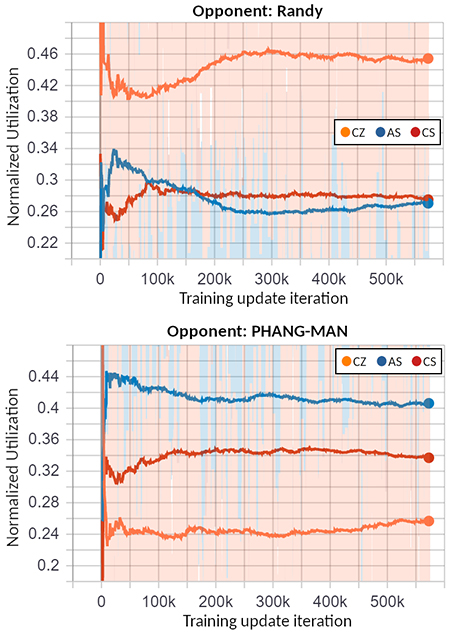}
\caption{Normalized utilization of low-level policies vs. PHANG-MAN (self-play) and Randy (random maneuver agent).}
\label{fig:LL_utilization}
\end{figure}

Percent utilization per episode of each low-level policy was calculated during policy selector training. When plotted over training episodes, unique usage profiles for different opponents are observed. As shown in Figure \ref{fig:LL_utilization}, the utilization profiles differ significantly when playing against an opponent that makes random actions every step, Randy (top), and an intelligent, aggressive agent such as itself (bottom).

During evaluation, we observed that the win rate of the hierarchical policy selector agent against any individual opponent was at least as great as the win rate of 
the highest performing low-level policy by itself. Additionally, across many of the opponents, the performance of the policy selector was significantly greater 
than the highest performing individual low-level policy. This indicates that the policy selector is able to effectively leverage the strengths of the low-level 
policies as well as generate unique and effective strategies by combining complementary behaviors. The superior performance of hierarchical policy selector agents over single-policy opponents can be seen in Table \ref{table:evolution} of the following subsection.

\subsection{Validation and Evolution of the Hierarchical Agents}

Using the training framework and techniques described in Section \ref{sec:training}, we trained and evaluated a variety of agents, and observed a progression of their abilities. In Table \ref{table:low-level}, we list the individual low-level policies used in the iterations of our hierarchical agent and provide a brief overview of each. 

In Table \ref{table:selector}, all iterations of our hierarchical agent are presented. The table details the reward function used to train the policy selector, the team members (low-level policies), and the initial conditions. These agents follow a chronological order and were composed by two or three of the low-level policies listed in Table \ref{table:low-level}. Each one of the agents was evaluated by playing 100 episodes against three unique opponents. The 100 episodes are the result of 50 unique scenarios played as both blue and red, and they have a near equal mixture of offensive, defensive, and neutral initial conditions, similar to the ones observed during the initial ADT events. The three opponents were chosen for their resemblance to previous iterations of competitors' agents and are a) Agent 1 from Table \ref{table:selector}, b) CS single-policy agent and c) AS single-policy agent. A summary of the evaluation results, with the total delta between wins and losses (Wins-Losses) across opponents, is shown in Table \ref{table:evolution}. Agent 9b was submitted to the ADT final event and was selected based on its superior Wins-Losses delta over the three representative evaluation opponents. Moreover, this agent was among the most aggressive (highest number of wins by gun snap) as well as among the most defensive (lowest number of losses) agents.
\begin{center}
\begin{table}[h]
\small
\begin{tabular}{ | c | p{6.8cm}  |}
\hline
\textbf{Policy} & \multicolumn{1}{|c|}{\textbf{Description}} \\
\hline
\multirow{6}{*}{CZ-1} &	Agent trained for ADT preliminary event with a reward function prioritizing attaining control-zone (Figure \ref{fig:CZ}) and action frequency of 10 Hz. \newline
\emph{Reward}: sum of the rewards given by Eqs. \eqref{eq:relative_position}-\eqref{eq:deck}. \newline
\emph{Initial conditions}: 50\% random, 50\% WEZ (1/2 offensive), with 25\% low-altitude bias. \\
\hline 
\multirow{4}{*}{CZ-2}  &	 Similar to CZ-1 agent but trained with action frequency of 50 Hz and additional too close penalty. \newline 
\emph{Reward}: sum of the rewards given by Eqs. \eqref{eq:relative_position}-\eqref{eq:too_close}. \newline
\emph{Initial conditions}: Same as CZ-1.\\
\hline
\multirow{6}{*}{CS}  &  Agent trained for ADT preliminary event with a reward function prioritizing conservative gun snaps (Figure \ref{fig:CS}) and action frequency of 50 Hz. \newline
\emph{Reward}: Eq. \eqref{eq:asps} with $\beta_B(d)=3$ and $\beta_R(d)=-3$. \newline
\emph{Initial conditions}: 100\% WEZ (1/2 offensive), with 25\% low-altitude bias. \\
\hline
\multirow{7}{*}{AS}	&  Agent trained with a reward function prioritizing aggressive gun snaps (Figure \ref{fig:CS}) and action frequency of 50 Hz. \newline
\emph{Reward}: Eq. \eqref{eq:asps} with $\beta_B(d)$ given by Eq. \eqref{eq:gsslopeblue} and $\beta_R(d)$ given by Eq. \eqref{eq:gsslopered}. \newline
\emph{Initial conditions}: 100\% WEZ (1/2 offensive), with 25\% low-altitude bias. \\
\hline
\end{tabular}
\caption{List of low-level policies used to compose a team.}
\label{table:low-level}
\end{table}
\end{center}

\begin{center}
\begin{table}[h]
\small
\begin{tabular}{ | c | p{6.8cm}  |}
\hline
\textbf{Agent} & \multicolumn{1}{|c|}{\textbf{Description}} \\
\hline
\multirow{4}{*}{1} &	 \textit{Reward}: $r_{{\phi}_{a}} + r_{gunsnap(blue)}+r_{gunsnap(red)}+r_{deck}$\newline
\textit{Initial conditions}: 25\% random, 25\% ADT, 50\% WEZ (1/2 offensive), with 25\% low-altitude bias. \newline
\textit{Team members}: CZ-1 and CS.\\
\hline
\multirow{2}{*}{2} &	 
Same as Agent 1 except: \newline
\textit{Team members}: CZ-2 (1 week of training) and CS. \\
\hline
\multirow{4}{*}{3} & Same as Agent 1 except: \newline
Historical states are encoded with a convolutional neural network. \newline
\textit{Team members}: CZ-2 (3 weeks of training) and AS. \\
\hline
\multirow{2}{*}{4} &	 Same as agent 3 except: \newline
\textit{Team members}: CZ-2 (3 weeks of training) and CS. \\
\hline
\multirow{2}{*}{5} &	 Same as agent 2 except: \newline
\textit{Team members}: CZ-2 (2 weeks training) and CS. \\
\hline
\multirow{5}{*}{6} &	 \textit{Reward}: Same as Agent 1. \newline
\textit{Initial conditions}: 25\% random, 25\% ADT (Gaussian noise added to avoid overfitting), 50\% WEZ (1/2 offensive), with 25\% low-altitude bias \newline
\textit{Team members}: CZ-2 (3 weeks of training) and AS. \\
\hline
\multirow{4}{*}{7} &	 \textit{Reward}: Eq. \eqref{eq:r_ps} \newline
\textit{Initial conditions}: Same as Agent 6. \newline
 \textit{Team members}: CZ-2 (3 weeks of training), CS and AS. \\
\hline
\multirow{2}{*}{8} &  Same as Agent 7 except: \newline
                      State and action values were truncated for the PS. \\
\hline
\multirow{5}{*}{9a,b} &  Same as Agent 8 except: \newline
                         State and action values were truncated for all low-level policies. \newline
\textit{Team members}: CZ-2 (3 weeks of training), CS and AS (a: 5 days of training, b: 6 days of training). \\
\hline
\multirow{4}{*}{10}	&
\textit{Reward}: $r_{gunsnap(blue)}+r_{gunsnap(red)}$ \newline
\textit{Initial conditions}: Same as Agent 6. \newline
\textit{Team members}: CZ-2 (4 weeks of training), CS and AS. \\
\hline
\multirow{6}{*}{11a,b}	&
\textit{Reward}: $r_{{\theta}_{t}} + r_{gunsnap(blue)}+r_{gunsnap(red)}+r_{deck}$ \newline 
\textit{Initial conditions}: 10\% random, 40\% ADT (with Gaussian noise), 50\% WEZ (1/2 offensive), with 25\% low-altitude bias. \newline
\textit{Team members}: Same as Agent 10.\newline
a: 3 days of training, b: 4 days of training \\
\hline
\end{tabular}
\captionof{table}{Details of hierarchical agent iterations.}
\label{table:selector}
\end{table}
\end{center}

\begin{table*}[t]
\makegapedcells
\small
\centering
\begin{tabular}{|c|c|c|c|c|c|c|c|c|c|c|c|c|c|}
\hline
\textbf{Agent} & \textbf{1} & \textbf{2} &            \textbf{3} & \textbf{4} & \textbf{5} & \textbf{6} & \textbf{7} &               \textbf{8} & \textbf{9a} & \textbf{9b} & \textbf{10} &      \textbf{11a} & \textbf{11b} \\
\hline
Number of Agents & 2 & 2 & 2 & 2 & 2 & 2 & 3 & 3 & 3 & 3 & 3 & 3 & 3 \\
\hline
\multicolumn{14}{l}{} \\
\multicolumn{14}{l}{\textbf{Opponent: Agent 1 from Table \ref{table:selector}}} \\
\hline
Total Wins & 58 & 56 & 44 & 57 & 51 & 56 & 54 & 58 & 60 & \textbf{64} & 58 & 59 & 61 \\
\hline
Wins by Gun Snap & 55 & 56 & 42 & 53 &  51 & 54 & 51 &  55 & 57 & \textbf{62} & 55 & 53 & 57 \\
\hline
Wins by Deck & 3 & 0 & 2 & 4 & 0 & 2 & 3 & 3 & 3 & 2 & 3 & \textbf{6} & 4 \\
\hline
Total Losses & 18 & 18 & 21 & 15 & 16 & 21 & 20 & 20 & 22 & 15 & 15 & \textbf{13} & 14 \\
\hline
Losses by Gun Snap & 18 & 18 & 21 & 15 & 16 & 20 & 20 & 20 & 22 & 15 & 15 & \textbf{13} & 14 \\
\hline
Losses by Deck & 0 & 0 & 0 & 0 & 0 & 1 & 0 & 0 & 0 & 0 & 0 & 0 & 0 \\                  
\hline                                                   
Wins-Losses & 40 & 38 & 23 & 42 & 35 & 35 & 34 & 38 & 38 & \textbf{49} & 43 & 46 &  47 \\
\hline
\multicolumn{14}{l}{} \\
\multicolumn{14}{l}{\textbf{Opponent: Conservative Shooter}} \\
\hline                                                                                                   
Total Wins & 49 & 46 & 42 & 53 & 52 & 53 & 60 & 58 & 58 & \textbf{62} & 38 & 59 &  61 \\
\hline
Wins by Gun Snap & 45 & 46 & 40 & 50 & 50 & 52 & \textbf{59} & 56 & 55 & 57 & 36 & 54 & 61 \\
\hline
Wins by Deck & 4 & 0 & 2 & 3 & 2 & 1 & 1 & 2 & 3 & \textbf{5} & 2 & \textbf{5} & 0\\
\hline   
Total Losses & 42 & 45 & 29 & 40 & 39 & 28 & 29 & 30 & 26 & 26 & 26 & 23 & \textbf{18} \\
\hline
Losses by Gun Snap & 42 & 42 & 29 & 39 & 37 & 28 & 28 & 29 & 25 & 26 & 24 & 23 & \textbf{16} \\
\hline
Losses by Deck & 0 & 3 & 0 & 1 & 2 & 0 & 1 & 1 & 1 & 0 & 2 & 0 & 2 \\
\hline
Wins-Losses & 7 & 1 & 13 & 13 & 13 & 25 & 31 & 28 & 32 & 36 & 12 & 36 & \textbf{43} \\
\hline
\multicolumn{14}{l}{} \\
\multicolumn{14}{l}{\textbf{Opponent: Aggressive Shooter}} \\
\hline
Total Wins & 50 & 56 & 50 & 49 & 61 & 60 & \textbf{67} & 60 & \textbf{67} & 65 & 54 & 61 & 62 \\
\hline
Wins by Gun Snap & 49 & 54 & 44 & 49 &  60 & 59 & \textbf{67} & 59 & 65 & 64 & 53 & 58 & 60 \\
\hline
Wins by Deck & 1 & 2 & \textbf{6} & 0 & 1 & 1 & 0 & 1 & 2 & 1 & 1 & 3 & 2 \\
\hline
Total Losses & 45 & 39 & 23 & 37 & 32 & 25 & 21 & 28 & 21 & \textbf{16} & 18 & 24 & 26 \\
\hline
Losses by Gun Snap & 40 & 35 & 23 & 36 & 31 & 24 & 21 & 28 & 19 & \textbf{16} & 17 & 24 & 25 \\
\hline
Losses by Deck & 5 & 4 & 0 & 1 & 1 & 1 & 0 & 0 & 2 & 0 & 1 & 0 & 1 \\
\hline
Wins-Losses & 5 & 17 & 27 & 12 & 29 & 35 & 46 & 32 & 46 & \textbf{49} & 36 & 37 &  36 \\
\hline
\multicolumn{14}{l}{} \\
\hline
\textbf{Total Wins-Losses} & 52 & 56 & 63 & 67 & 77 & 95 & 111 &  98 & 116 & \textbf{134} & 91 & 119 & 126 \\
\hline
\end{tabular}
\caption{Evaluation of the hierarchical agents listed in Table \ref{table:selector}.}
\label{table:evolution}
\end{table*}

\subsection{PHANG-MAN at the ADT Final Event}

Ultimately, the final test of our agent was performed live at the ADT final event which consisted of 3 days of competition. On Day 1, competitors faced agents developed by JHU-APL in collaboration with DARPA. Then on Day 2,
all competitors played each other in a best of 20 round robin tournament, with starting conditions ranging from offensive to defensive. On Day 3, the four 
teams with the highest cumulative score competed in a single elimination championship tournament. In each round of the championship tournament, teams played 20 rounds with neutral starting conditions. These teams also had the opportunity to play against a graduate of the USAF F-16 Weapons Instructor Course in a best of 5 match. 

Our agent finished the initial 2 days of the competition in $2^{nd}$ place and qualified for the championship tournament. On Day 3, PHANG-MAN won its semi-final round and was defeated in the final round by Heron Systems' agent, finishing the competition in $2^{nd}$ place overall. Finally, our agent emerged victorious (5 W - 0 L) from its match with the USAF Weapons Instructor pilot. The match-ups were characterized by PHANG-MAN taking aggressive shots from head on and the side, while also capitalizing on any mistakes that the human pilots made that gave up their control zone. Overall, PHANG-MAN demonstrated strategic use of aggressive and conservative tactics, reflecting its hierarchical architecture. A full video coverage of the final event can be found online\footnote{https://www.youtube.com/watch?v=NzdhIA2S35w}, and a more detailed description is provided in supplemental material.

\section{Discussion}

\subsection{Low-level policies}
Unlike traditional hierarchical RL architectures like HIRO \cite{nachum}, in which low-level policies execute sub-tasks, in our approach low-level policies were able to execute the entire task (i.e., defeating the adversary). In Frans et al. \cite{frans2018}, low-level policies were used to solve parts of the task, while in our application, low-level policies were selected as a way to optimally adapt to the opponent's maneuvers through the use of different reward functions (Section \ref{subsec:LL_policies}). Although a more extensive analysis of these policies would be interesting, in this work we focused on the final hierarchical agent and its performance.

\subsection{Policy selector}
As described in Section \ref{subsec:effective_selector}, the policy selector was able to strategically select low-level policies according to its positional advantage (Figure \ref{fig:LL_selection}) and opponent type (Figure \ref{fig:LL_utilization}). Similar to previous work in which different skills are selected at different stages of the task \cite{frans2018,Lee2019composing}, our high-level policy selected the low-level policy according to the current context of the engagement. 

In this work, we employed a curriculum learning strategy (Section \ref{subsec:training_strategy}) which selects adversaries according to their performance metrics (Section \ref{subsec:performance_metrics}), and achieved progressive improvement as demonstrated in Table \ref{table:evolution}. However, a multi-agent training perspective \cite{Sun2021} and a more game theoretic approach for population-level training \cite{Garnelo2021} are promising techniques to be explored.

\subsection{Limitations and Future Work}
The one-year timeline of ADT provided many new avenues of future research in this problem space. To this end Lockheed Martin is making significant investments 
to further research and develop this work.
Alongside deeper algorithmic research specific to the air-to-air domain we are adapting our current efforts with the goal of full-scale deployment. Enhancing 
the simulation environment is the first step in this direction: modification and additions to the observation space (e.g., incorporating domain randomization 
techniques \cite{tobin2017domain}, imperfect information, partial and estimated knowledge) and incorporating different platform aerodynamics models 
(quad-copter, small fixed-wing unmanned aerial vehicles, F-22, among others). Narrowing the gap between simulation and real-world deployment (sim2real) is critical to 
rapid-prototyping and validation, as shown by several recent studies \cite{Sunderhauf2018,Zhao2020,Nachum2019,Hanna2021,Ranaweera2021}. These enhancements will enable
smoother transition as we deploy on sub-scale and eventually full-scale aircraft.

\section{Conclusion}
The AlphaDogfight Trials sought to challenge competitors to develop high performing agents able to excel in air-to-air combat. We developed a hierarchical agent 
that succeeded in competition, ranking $2^{nd}$ place overall and defeating a graduate of the USAF Weapons School's F-16 Weapons Instructor Course. LM intends 
to continue investing in this research and fully explore the potential of these algorithms and their path to deployment.
The main conclusions of this work can be summarized as: (i) hierarchical RL, along with modern auxiliary techniques, can be successfully applied to air combat scenarios in high-fidelity environments, producing highly effective maneuvers and tactics; (ii) reward shaping of low-level policies provides a mechanism to incorporate expert pilot knowledge; (iii) training framework design choices play a crucial role in the development of a high performing RL agent. 

\section*{Acknowledgment}
The authors would like to thank DARPA and JHU-APL for being great hosts to the program. The authors would like to thank the members of the research and development team who made impactful contributions to the project: M. Tarascio (former VP of Artificial Intelligence at Lockheed Martin), W. Xie, J. Stanco, B. Liston, J. “Vandal” Garrison, and P. Miskech.

Any opinions, findings, conclusions or recommendations expressed are those of the authors and do not necessarily reflect the views of DARPA, the United States Air Force, the United States Department of Defense, or the United States Government.

\nocite{ide}

\bibliographystyle{IEEEtran}
\bibliography{HDRL_ADT}

\begin{IEEEbiography}[{\includegraphics[width=1in,height=1.25in,clip,keepaspectratio]{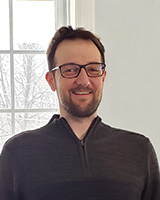}}]
    {Adrian P. Pope}{\space} received the B.S. degree in physics from the University of California, Santa Barbara and the M.S. degree in electrical engineering from San Jose State University. He is the co-founder of Primordial Labs, a company focused on developing human-centric autonomy. As a researcher engineer at Lockheed Martin from 2015 to 2021, he served as the principal investigator on several Science and Technology programs. His current work is focused on autonomous control of unmanned systems and human-machine teaming.
\end{IEEEbiography}

\begin{IEEEbiography}[{\includegraphics[width=1in,height=1.25in,clip,keepaspectratio]{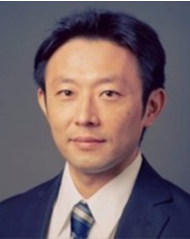}}]
    {Jaime S. Ide}{\space}received the B.S. degree in mechatronic engineering, and the M.S. and the Ph.D. degrees in engineering (artificial intelligence) from the University of Sao Paulo, Brazil. From 2007-2010, he worked on computational and Bayesian methods applied to neuroimaging at the University of Pennsylvania and Yale University. From 2010 to 2015, he was an Assistant Professor at the Federal University of Sao Paulo, Brazil. From 2016-2019, he worked as an Associate Research Scientist at Yale University School of Medicine. Since 2019, he has been a Research Engineer with the Lockheed Martin Artificial Intelligence Center, where he develops machine learning in unmanned aircraft and cognitive systems. Dr. Ide was the recipient of the Sao Paulo Research Foundation FAPESP’s Young Investigators Grants in 2011.
\end{IEEEbiography}

\begin{IEEEbiographynophoto}
{Daria Mi\'{c}ovi\'{c}} received the B.S. degree in mathematics from the University of Washington. She worked as a Machine Learning Engineer at the Lockheed Martin Artificial Intelligence Center.
\end{IEEEbiographynophoto}

\begin{IEEEbiography}
[{\includegraphics[width=1in,height=1.25in,clip,keepaspectratio]{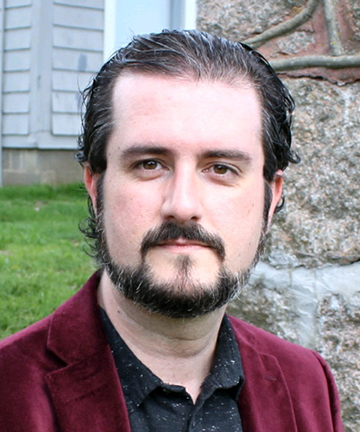}}] 
	{Henry Diaz}{\space} received the B.S. degree in computer engineering from Florida International University. He is an associate LM Fellow within the Lockheed Martin Artificial Intelligence Center. His work focuses on rapid prototyping and the integration and deployment of machine learning algorithms on size, weight, and power constrained systems.
\end{IEEEbiography}

\begin{IEEEbiography}[{\includegraphics[width=1in,height=1.25in,clip,keepaspectratio]{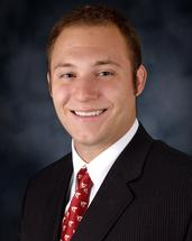}}]
    {Jason C. Twedt}{\space} received the B.S. degree in physics from Mansfield University of Pennsylvania and the M.S. from the Virginia Polytechnic Institute and State 
    University in aerospace engineering. Currently, he is the manager of the Computer Vision Systems group at the Lockheed Martin Artificial Intelligence Center, where he leads 
    a team focused on building capabilities that leverage advancements in deep learning and computer vision to create advanced perception capabilities for one or 
    more sensing agents.
\end{IEEEbiography}

\begin{IEEEbiography}[{\includegraphics[width=1in,height=1.25in,clip,keepaspectratio]{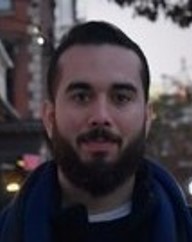}}]
    {Kevin Alcedo}{\space} received both the B.S. degree in Mechanical Engineering and Applied Mechanics and the M.S. degree in Robotics from the University of Pennsylvania. 
    Currently, he is an Engineering Research Manager at the Lockheed Martin Artificial Intelligence Center where he leads a team focused on the R\&D of intelligent agent capabilities. Previously, he was a Machine Learning Engineer at NBC Universal where he developed measurement and optimization algorithms used across the company’s entertainment portfolio. He holds various patents in the field of ``Soft Robotics''. 
\end{IEEEbiography}

\begin{IEEEbiography}[{\includegraphics[width=1in,height=1.25in,clip,keepaspectratio]{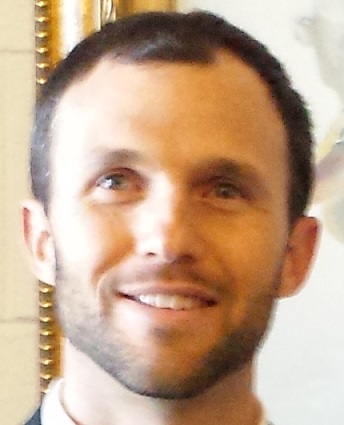}}]
    {Thayne T. Walker}{\space}received the B.S. degree in computer science from Southern Utah University, M.S. in computer science from Colorado State University and the Ph.D. in artificial intelligence at University of Denver. He has background in electronic warfare, multi-source data fusion, cloud computing systems, combinatoric optimization and multi-agent systems. He leads several research and development projects at Lockheed Martin and has published novel algorithms for multi-agent pathfinding. 
\end{IEEEbiography}

\begin{IEEEbiography}[{\includegraphics[width=1in,height=1.25in,clip,keepaspectratio]{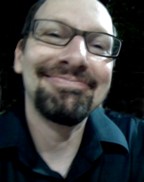}}]
    {David Rosenbluth}{\space} received the B.S. in mathematics and theoretical computer science from Columbia University and the Ph.D. in computation and neural systems from Caltech. He is a LM Fellow within the Lockheed Martin Artificial Intelligence Center and has 25 years of experience in this domain. The current focus of his research is machine learning generative models, neuromorphic systems, and integration of machine learning and machine reasoning. He has published research on the topics of electrophysiology and visual neuroscience, neuromorphic photonics, machine learning, and mathematics.
\end{IEEEbiography}

\begin{IEEEbiography}[{\includegraphics[width=1in,height=1.25in,clip,keepaspectratio]{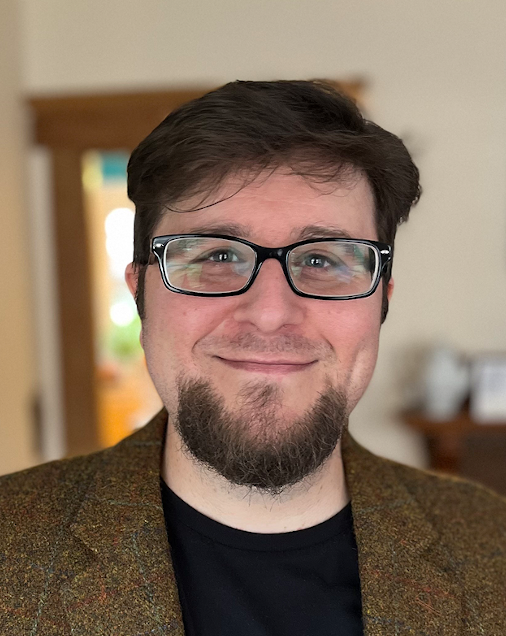}}]
    {Lee Ritholtz}{\space} received the B.S. and M.S. degrees in electrical engineering from the State University of New York at Binghamton. He is the co-founder of Primordial Labs, a company focused on developing human-centric autonomy. He was previously the Director and Chief Architect for the Applied AI organization at the Lockheed Martin Artificial Intelligence Center. His current work focuses on machine learning for autonomous decision making, sensor processing at the edge, and human-machine interfaces for unmanned systems.
\end{IEEEbiography}

\begin{IEEEbiography}[{\includegraphics[width=1in,height=1.25in,clip,keepaspectratio]{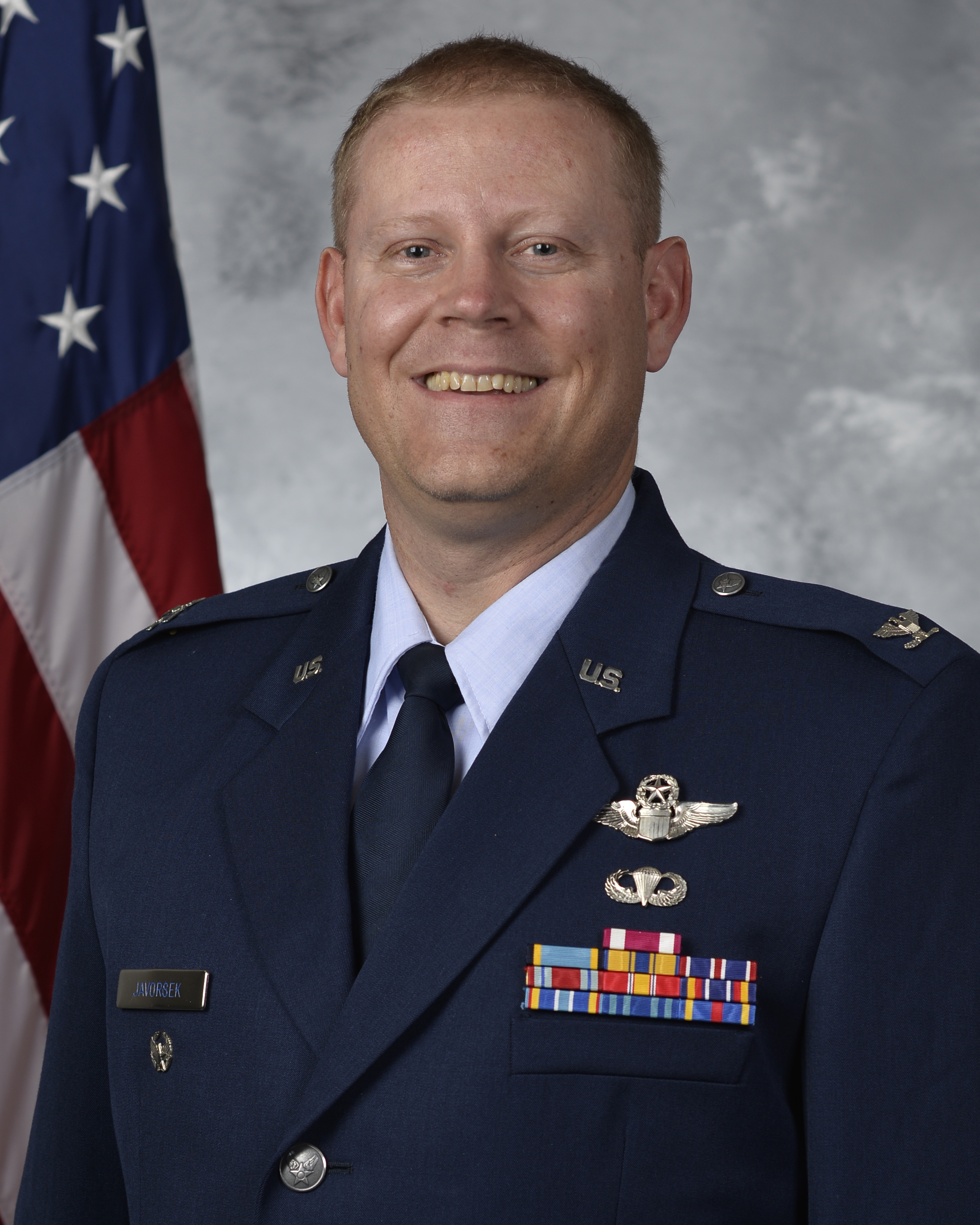}}]
       {Daniel Javorsek II}{\space} holds several advanced degrees including the Ph.D. in physics from Purdue University. He is a former Program Manager for DARPA and is currently the Commander of Detachment 6, Air Force Operational Test and Evaluation Center (AFOTEC), Nellis Air Force Base, and Director, F-35 U.S. Operational Test Team. As the Commander of AFOTECs Detachment 6 he leads the planning, execution, and reporting on realistic, objective, and impartial operational test and evaluation of fighter aircraft.
\end{IEEEbiography}

\clearpage
\onecolumn
\thispagestyle{empty}
\vspace*{2cm}
\begin{center}
\begin{Large}
\textbf{SUPPLEMENTARY MATERIAL}
\end{Large}

\vspace{20mm}

\begin{LARGE}
Hierarchical Reinforcement Learning for Air Combat at DARPA's AlphaDogfight Trials
\end{LARGE}

\vspace{20mm}

Adrian P. Pope, Jaime S. Ide, Daria Mi\'{c}ovi\'{c}, Henry Diaz, Jason C. Twedt, Kevin Alcedo,\\
Thayne T. Walker, David Rosenbluth, Lee Ritholtz, and Daniel Javorsek II
\end{center}

\newpage
\setcounter{section}{0}
\setcounter{figure}{0}
\setcounter{table}{0}
\renewcommand{\thefigure}{S\arabic{figure}}
\renewcommand{\thetable}{S\arabic{table}}
\renewcommand{\thesection}{S\arabic{section}}
\section{PHANG-MAN at the ADT Final Event}
The ADT final event consisted of 3 days of competition. On Day 1, competitors faced agents developed by JHU-APL in collaboration with DARPA. Then on Day 2,
all competitors played each other in a best of 20 round robin tournament, with starting conditions ranging from offensive to defensive. On Day 3, the four 
teams with the highest cumulative score competed in a single elimination championship tournament. In each round of the championship tournament, teams 
played 20 rounds with neutral starting conditions. These teams also had the opportunity to play against a graduate of the USAF F-16 Weapons Instructor 
Course in a best of 5 match.

\begin{figure}[h]
\centering
\includegraphics[width=0.5\columnwidth]{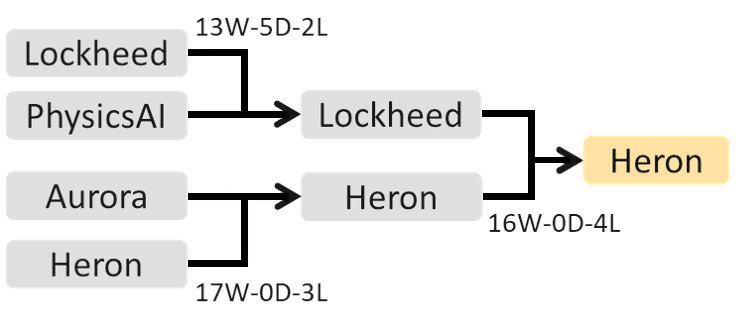}
\caption{Day 3 tournament results.}
\label{fig:day3results}
\end{figure}

Our agent, the submission from Lockheed Martin, finished the initial 2 days of the competition in $2^{nd}$ place and qualified for the championship tournament. On Day 3, PHANG-MAN won its 
semi-final round and was defeated in the final round by Heron Systems' agent, finishing the competition in $2^{nd}$ place overall. The tournament bracket 
and results are shown in Figure \ref{fig:day3results}. The semi-finals and finals can be viewed online.\footnote{https://www.youtube.com/watch?v=NzdhIA2S35w}

In the championship round, we faced a very aggressive and highly accurate adversary. In most cases, PHANG-MAN did not survive the initial exchange of 
nose-to-nose gun snaps and matches would end with the opponent having a small amount of health remaining. Although we scored 7\% more total shots 
against our opponent, our average gun snap was from a greater distance and as a result the average damage per gun snap was lower. Our agent appeared to 
disengage its offense inside of 800 feet in favor of better positioning for the next exchange, while the opposing agent would continue to aggressively pursue 
head on. Whenever PHANG-MAN survived the initial exchange, it was generally able to achieve and maintain an advantageous position.

We suspect that this bias toward future positioning over immediate scoring is the result of a strategy we used during training in which we increased the
health of the agents by a factor of ten. Additionally, by not providing the opponent's health as a state value or a reward proportional to the opponent's 
health, we inadvertently trained our agent to allow giving away near victories.

\begin{figure}[h]
\centering
\includegraphics[width=.5\columnwidth]{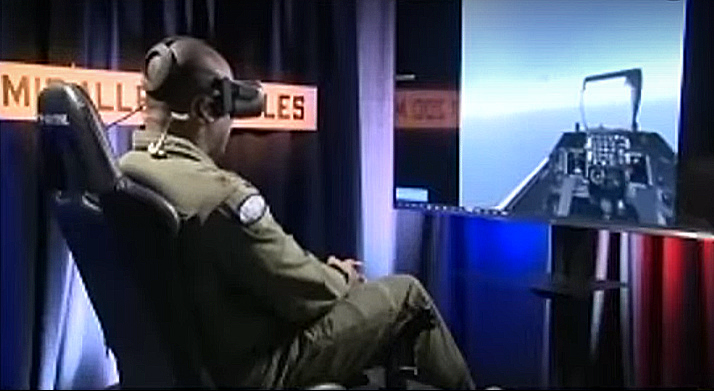}
\caption{VR enabled cockpit for human vs. AI matchups.}
\label{fig:VR_cockpit}
\end{figure}

Overall, PHANG-MAN demonstrated strategic use of aggressive and conservative tactics. We believe it has the potential to be a trustworthy co-pilot or wingman 
that can effectively execute real-world tactics in situations where self-preservation is paramount and aggressive nose-to-nose are not feasible.

For the matches between the top competitors and the human USAF pilot, a high-fidelity VR enabled cockpit was provided by the DARPA and JHU-APL 
team (Figure \ref{fig:VR_cockpit}). This allowed the human pilot to visually track the opponent as he typically would in a real engagement. A dashboard was displayed for the pilot providing a simplified view of pertinent information (e.g track angle, relative distance to opponent, altitude, fuel, etc). In order to 
provide an additional visual assist for the pilot, an icon pointed out the direction to the opponent when out of view and a red flashing overlay was 
shown when receiving a gun snap. 

PHANG-MAN emerged victorious (5 W - 0 L) from its match with the USAF Weapons Instructor pilot. The match-ups were characterized 
by PHANG-MAN taking aggressive shots from head on and the side, while also capitalizing on any mistakes that the human pilot made that gave up their control zone.

\section{Training Framework}
For this effort, a custom off-policy distributed training framework was developed in Pytorch (v1.3.1). The framework is similar to the Ape-X architecture but enhanced to accommodate SAC. It is composed of distributed actors, a central learner, and a central prioritized experience replay (PER) buffer. Each actor has its own instance of the environment and a copy of the most recent policy weights. The centralized PER buffer managed the most significant experiences, while the central learner continuously updated the policy and value networks.

\begin{figure}[h!]
\centering
\includegraphics[width=.75\columnwidth]{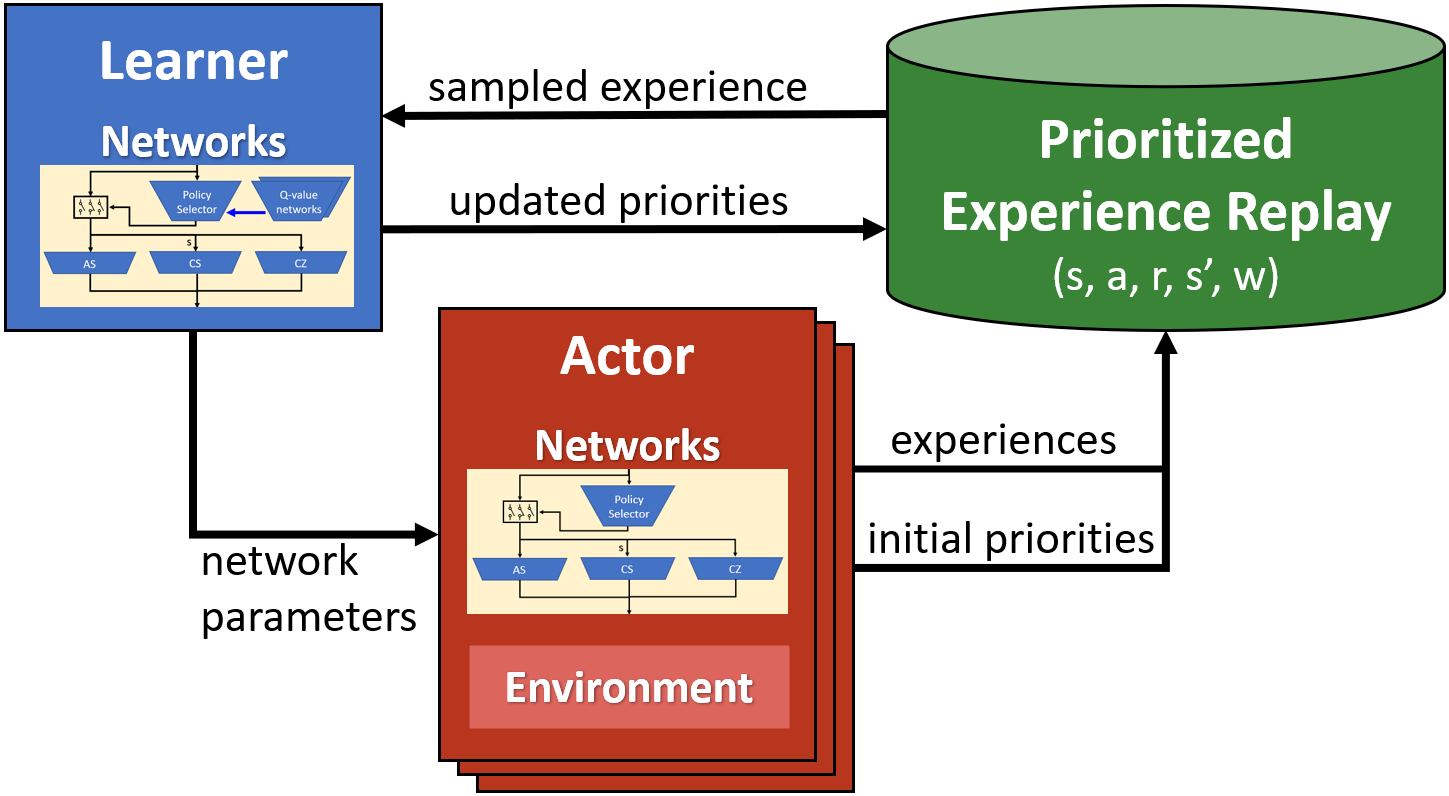} 
\caption{High-level architecture diagram of the distributed training framework.}
\label{fig:apex}
\end{figure}

\clearpage
\section{State Space}
\begin{table}[h!]
\small
\centering
\begin{tabular}{|c|c|c|c|c|}
\hline
\textbf{State} & \underline{\textbf{Own}} & \textbf{Opp} & \textbf{Delta} & \textbf{Delta Opp} \\
\hline
altitude (ft) & LLP / PS & LLP / PS & LLP / PS & LLP / PS \\
\hline 
roll (rad) & LLP / PS & LLP / PS & & \\
\hline
pitch (rad) & LLP / PS & LLP / PS & & \\
\hline
heading (deg) & LLP / PS & LLP / PS & & \\
\hline
bodyframe x-axis velocity (ft/s) & LLP / PS & LLP / PS & & \\
\hline
bodyframe y-axis velocity (ft/s) & LLP / PS & LLP / PS & & \\
\hline
bodyframe z-axis velocity (ft/s) & LLP / PS & LLP / PS & & \\
\hline
roll rate (rad/s) & LLP / PS & LLP / PS & &  \\
\hline
pitch rate (rad/s) & LLP / PS & LLP / PS & & \\
\hline
yaw rate (rad/s) & LLP / PS & LLP / PS & & \\
\hline
bodyframe x-axis acceleration (ft/s$^2$) & LLP & & & \\
\hline
bodyframe y-axis acceleration (ft/s$^2$) & LLP & & & \\
\hline
bodyframe z-axis acceleration (ft/s$^2$) & LLP & & & \\
\hline
roll acceleration (rad/s$^2$) & LLP & & & \\
\hline
pitch acceleration (rad/s$^2$) & LLP & & & \\
\hline
yaw acceleration (rad/s$^2$) & LLP & & &\\
\hline
alpha: angle of attack (deg) & LLP & LLP & LLP & LLP \\
\hline
beta: sideslip (deg) & LLP & LLP & LLP & LLP \\
\hline
normalized left aileron position & LLP & & LLP &\\
\hline
normalized right aileron position & & & & \\
\hline
normalized left flaperon position & & & & \\
\hline
normalized right flaperon position & & & & \\
\hline
normalized speedbrake position & & & & \\
\hline
normalized trailing edge flap position & LLP & & LLP & \\
\hline 
normalized leading edge flap position & LLP & & LLP &  \\
\hline
left differential horizontal tail (rad) & LLP & & LLP &  \\
\hline
right differential horizontal tail (rad) & LLP & & LLP & \\
\hline
normalized rudder position & LLP & & LLP &  \\
\hline
normalized throttle position & LLP & & LLP &  \\
\hline
fuel in tanks (lbs) & LLP & & & \\
\hline
simulation time (s) & & & &  \\
\hline
engine thrust (lbs) & LLP & & LLP &  \\
\hline
x position (E) relative to scenario center (ft) & & & &\\
\hline
y position (N) relative to scenario center (ft) & & & & \\
\hline
z position (U) relative to scenario center (ft) & & & &\\
\hline
distance to opponent (ft) & LLP / PS & & LLP / PS & \\
\hline
track angle (rad) & LLP / PS & LLP / PS & LLP & LLP \\
\hline
height above altitude hard deck (ft) & & & & \\
\hline
aircraft health & & & & \\
\hline
whether opponent is in the WEZ & & & & \\
\hline 
*airspeed (ft/s) & LLP / PS & LLP / PS & LLP / PS & LLP / PS \\
\hline
*bearing to opponent, accounting for roll/pitch (deg) & LLP / PS & LLP / PS & LLP / PS & LLP / PS \\
\hline
*elevation to opponent, accounting for roll/pitch (deg) & LLP / PS & LLP / PS & LLP / PS & LLP / PS \\
\hline
*bearing to opponent, plane as point mass (deg) & LLP / PS & LLP / PS & LLP / PS & LLP / PS \\
\hline
*elevation to opponent, plane as point mass (deg) & LLP / PS & LLP / PS & LLP / PS & LLP / PS \\
\hline
\end{tabular}
\caption{State space of the environment and quantities used to train the low-level policies (LLP) and policy selector (PS). The Own column includes values of agent itself, the Opp column includes values of the agent's opponent, and the Delta / Delta Opp columns include the change in Own / Opp values over the previous 10 steps. States with an asterisk are derived from the states provided by the simulation environment.}
\label{table:states}
\end{table}

\end{document}